\documentclass{article}

\usepackage{arxiv} 
\usepackage{graphicx} 
\usepackage[sort,numbers,compress]{natbib} 
\usepackage{amsmath}
\usepackage{amssymb}
\usepackage{xcolor}
\usepackage{placeins}
\usepackage{subcaption}
\usepackage{tikz}
\usetikzlibrary{positioning, fit, arrows.meta, calc}
\usepackage{hyperref}
\usepackage{cleveref}
\usepackage{enumitem}
\usepackage{booktabs}
\usepackage{sidecap}
\usepackage{microtype}
\sidecaptionvpos{table}{c}
\sidecaptionvpos{figure}{c}
\usepackage{caption}
\usepackage[T1]{fontenc}
\newcommand{\node}{\boldsymbol{v}}

\definecolor{darkspringgreen}{rgb}{0.09, 0.45, 0.27}

\newcommand{\method}{\textsc{Decide}}

\definecolor{mvfocus}{HTML}{D55E00} 
\definecolor{mvseta}{HTML}{0072B2}  
\definecolor{mvsetb}{HTML}{009E73}  
\definecolor{mvsetc}{HTML}{CC79A7}  
\definecolor{mvsetd}{HTML}{E69F00}  

\newcommand{\focustext}[1]{\textcolor{mvfocus}{#1}}

\tikzset{
  tnode/.style={
    rectangle, draw=black!65, thick,
    minimum size=5.5mm, inner sep=1pt,
    font=\scriptsize\bfseries
  },
  tnodehigh/.style={
    rectangle, draw=mvfocus, fill=mvfocus!20, line width=1.6pt,
    minimum size=5.5mm, inner sep=1pt,
    font=\scriptsize\bfseries
  },
  lnode1/.style={draw=mvseta, line width=1.6pt},
  lnode2/.style={draw=mvsetb, line width=1.6pt},
  lnode3/.style={draw=mvsetc, line width=1.6pt},
  lnode4/.style={draw=mvsetd, line width=1.6pt},
  setupnode/.style={rectangle, draw=black!65, thick,
    minimum width=10mm, minimum height=5mm, inner sep=1pt,
    font=\small\bfseries},
  setupleaf/.style={setupnode, minimum width=6mm},
  setupnodehigh/.style={rectangle, draw=mvfocus, fill=mvfocus!20,
    line width=1.6pt,
    minimum width=10mm, minimum height=5mm, inner sep=1pt,
    font=\small\bfseries},
  highedge/.style={draw=mvfocus, line width=1.2pt, -latex},
  mergednode/.style={setupnode, fill=black!8, densely dashed},
  noopleaf/.style={setupleaf, draw=black!35, text=black!35, fill=white},
  cfleaf/.style={setupleaf, densely dashed, fill=white},
  matchedarrow/.style={-latex, densely dotted, thin, black!60},
}

\definecolor{lavender}{rgb}{0.55, 0.40, 0.75}

\begin{document}

\title{Pinpointing Pivot Points: The DECIDE Framework for Data Analysis Decision Importances}

\author{%
  \parbox{\dimexpr\linewidth-2\tabcolsep\relax}{\centering
    \mbox{Zachary T. Rewolinski$^{1,\dagger}$},
    \mbox{Jan Simson$^{2,3,4,\dagger}$},
    \mbox{Frauke Kreuter$^{2,3}$},
    \mbox{Christoph Kern$^{2,3,4,\ddagger}$},
    \mbox{Bin Yu$^{1,\ddagger}$}
    \\[4pt]
    {\normalfont\small
      \mbox{$^{1}$UC Berkeley} \quad
      \mbox{$^{2}$LMU Munich} \quad
      \mbox{$^{3}$Munich Center for Machine Learning (MCML)} \quad
      \mbox{$^{4}$University of Mannheim}
    }
  }%
}

\maketitle
\begin{NoHyper}
\renewcommand{\thefootnote}{}%
\footnotetext{$^{\dagger}$Equal contribution. \quad $^{\ddagger}$Equal advising.}%
\end{NoHyper}

\begin{abstract}

    %
    Human judgment calls made throughout the data science life cycle, including analytical decisions, can change the conclusions drawn from data in both research and industry.
    Independent analysts often make different choices when given the same domain questions and the same data, leading to often unacceptable variations in results and contributing to the ongoing reproducibility crisis across scientific disciplines and beyond.
    The rapid adoption of AI and agentic tools threatens to amplify this crisis by reducing the transparency of the decision-making process.
    Several approaches have been developed to explicitly address this variation by following the same basic strategy: repeating an analysis under different combinations of analytical decisions and inspecting the distribution of outcomes. Examples include multiverse analysis, specification curve analysis, and vibration of effects.
    More broadly, the Predictability-Computability-Stability (PCS) framework for veridical data science assesses how documented judgment calls across the data science life cycle affect conclusions, with predictability serving as a reality-check that a model captures signal.
    Each of these approaches results in a ``multiverse'' object that represents a structured collection of analyses indexed by analytical decisions.
    While the distribution of outcomes reflects the overall variability of conclusions, estimating the influence of each individual decision requires additional tools and methods.
    A common heuristic is to train a model to predict the outcome from these decisions and treat its coefficients or feature importances as proxies for each decision's influence.
    Although this strategy may work in certain cases, it assumes the model is correctly specified and accurately predicts the outcome, which we show is not always the case.
    In this work, we introduce \method\ (\textbf{D}ecision \textbf{E}valuation by \textbf{C}alculating \textbf{I}mportance along \textbf{D}iverging \textbf{E}dges), a model-free framework that quantifies each decision's importance
    directly from the actual outcomes across the plausible decision space, avoiding potential issues of predictability by design.
    %
    %
    We demonstrate the effectiveness of \method\ by applying it to both well-known and real-world multiverse analyses.
    We further provide what is, to our knowledge, the first formal definition of the multiverse, mathematically grounding the framework and enabling future methodological research on multiverse analysis{\renewcommand{\thefootnote}{\fnsymbol{footnote}}\footnote{Code is available at \url{https://github.com/reliable-ai/decide_analysis}.}}\setcounter{footnote}{0}.
\end{abstract}

\section{Introduction}
\label{sec:introduction}

%
While conclusions drawn from data are often treated as infallible, analysts answering the same research question using the same dataset can make different methodological choices and arrive at conflicting results \citep{silberzahn2015crowdsourced,silberzahn2018manya}.
This calls into question the robustness of empirical findings.
A striking illustration of this is a recent study by Aczel et al.~\cite{aczel2026investigating}, in which $100$ published social science findings were independently reanalyzed at least five times by analysts using the original data.
Only $34\%$ of the independent reanalyses closely matched the original result, with $26\%$ either finding no effect or reaching the opposite conclusion.
This sensitivity of results to analytical choices is not a new observation \cite{yu2013stability, gelman2014statistical, breznau2022observing}, nor is it unique to the social sciences, as the natural sciences face the same challenge \cite{gould2025same}.
Furthermore, the rapid adoption of agentic programming tools threatens to exacerbate this issue by introducing methodological decisions that analysts accept without full scrutiny.
Conventional statistical inference largely ignores this source of variability, leaving practitioners ill-equipped to measure and explain it, thus contributing to overconfident conclusions and irreproducible findings.
Understanding the drivers of this variation is crucial not only for achieving accurate findings, but also for informing real-world decisions that rely on these findings.

To address this unacknowledged uncertainty and combat the reproducibility crisis, researchers have developed several parallel lines of work over the past decade.
%
The Predictability-Computability-Stability (PCS) framework~\citep{yu2013stability, yu2020veridical, yu2024veridical, rewolinski2026pcs} applies across the entire data science life cycle, from data collection to communicating results.
It requires conclusions to pass a predictability ``reality-check'' and to be stable under reasonable perturbations to judgment calls at each step.
PCS calls for these judgment calls and their perturbations to be documented in domain context and satisfy computational resource constraints.
Independently, multiverse analysis~\citep{steegen2016increasing}, specification curve analysis~\cite{simonsohn2020specification}, and vibration of effects~\cite{patel2015assessment} enumerate ``reasonable'' options for each analytical decision to examine a combinatorial set of plausible data analyses and mostly differ in how they analyze or visualize the results of this set.
Multiverse analysis~\cite{steegen2016increasing} has emerged as the predominant framework for exploring the consistency of results and has been proposed as the umbrella term for enumeration-based approaches~\citep{short2026multicurious,young2024multiverse}.
%
%
By computing the outcome of every plausible analysis, practitioners can observe the entire distribution of potential results.
%
%
While the multiverse contains the exact structural information needed to explain analytical instability, current methods fail to fully exploit this, an issue explicitly highlighted by Short et al.~\citep{short2026multicurious}.
A multiverse with highly variable outcomes tells an analyst that something is fragile, but additional tools are necessary to infer \textit{which particular decisions} are responsible.
This leaves practitioners without the targeted guidance needed to understand the root causes.
Our work seeks to bridge this gap by providing a principled method for pinpointing the specific decisions that drive analytical variability.
Specifically, we combine the strengths of PCS and multiverse analysis to directly analyze the stability of a multiverse with respect to each individual decision.

To distill actionable insights from the descriptive complexity of a multiverse, savvy researchers may fit a variety of different surrogate models (e.g., linear regression, random forest, etc.) to the multiverse output and interpret the model's feature importances as proxies for decision importances~\citep{young2025analytical,short2026multicurious}.
Whether or not these models fit the data is seldom reported and, as we demonstrate in Section \ref{sec:model-dependent}, these surrogate models often fit the multiverse poorly.
Consequently, practitioners should be highly cautious in using surrogate models, as the relationships derived from them are frequently unfaithful to the true dynamics of the analysis.
Global sensitivity analysis (GSA), which quantifies the contributions of each input to the observed variation in the output of a system, presents a principled alternative.
Recent work has applied variance-based GSA to multiverse analysis~\cite{saltelli2025global}.
However, the structure of analytical decisions in multiverses is different from the setting standard GSA methods were designed for.
In particular, decisions are categorical and can be restricted or branch-dependent.
Furthermore, analysts may be interested in quantities other than average-case importance, such as how a decision can influence the tail of an outcome distribution.

In this work, we introduce the \method\ (\textbf{D}ecision \textbf{E}valuation by \textbf{C}alculating \textbf{I}mportance along \textbf{D}iverging \textbf{E}dges) framework to provide an alternative approach to estimating decision importance.
By directly evaluating the structure of the multiverse rather than fitting an intermediate surrogate model, our framework avoids the pitfalls of proxy-based approaches and faithfully measures the influence of analytical judgment calls.
This model-free design yields a more robust interpretation of decision importance, achieving actionable insights without sacrificing the full coverage provided by the multiverse.
\method\ additionally connects to GSA by recovering Sobol' indices as a special case, while its flexible framework handles restricted and branch-dependent decisions as well as risk profiles beyond the average case.
In summary, we make four core contributions:
\begin{itemize}[
    topsep=0pt,
    partopsep=0pt,
    itemsep=1pt,
    parsep=0pt
]
    \item A formal mathematical definition and geometric representation of multiverse analyses.
    \item The \method\ framework, which systematically quantifies the influence of analytical decisions and their interactions in a flexible, model-free manner.
    \item An empirical evaluation across five real-world multiverse analyses demonstrating that the \method\ framework yields stable and computationally efficient decision importance estimates.
    \item A synthetic example demonstrating the framework's flexibility through a worst-case analysis, in which \method\ flags decisions which are benign on average but introduce severe vulnerabilities in rare configurations.
\end{itemize}

\section{Related Work}
\label{sec:related-work}


\subsection{Researcher Degrees of Freedom}

Empirical research often requires analysts to navigate a series of discretionary choices on topics including data cleaning, modeling, experimental design, and more.
These ``researcher degrees of freedom'' \citep{simmons2011false} generate a ``garden of forking paths'' \citep{gelman2014statistical} where different defensible choices yield pipelines with distinct conclusions.
Crowdsourced multi-analyst studies demonstrate how independent teams using the same data can arrive at opposing results \citep{silberzahn2015crowdsourced,silberzahn2018manya,breznau2022observing,coretta2023multidimensional}.
This variability introduces an additional layer of uncertainty into published conclusions \citep{breznau2022observing}, contributing to the broader reproducibility crisis that spans not only the social and natural sciences, but machine learning (ML) and artificial intelligence (AI) as well \citep{ioannidis2005why,begley2012replication,laraway2019overview,hutson2018artificiala,semmelrock2023reproducibility,gundersen2018state,herrmann2024position}. 
Addressing this source of variation requires observing the space of decisions instead of reporting a single pipeline \citep{yu2013stability, steegen2016increasing}.


\subsection{Stability Analysis}

To measure this decision-based uncertainty, several methodological frameworks have emerged.
Steegen et al.~\citep{steegen2016increasing} propose multiverse analysis, an approach that systematically executes all plausible combinations of data preprocessing and modeling choices.
Complementing this strategy, Simonsohn et al.~\citep{simonsohn2020specification} introduce specification curve analysis to display and evaluate results across reasonable analytical specifications, while Patel et al.~\citep{patel2015assessment} propose measuring the vibration of effects to quantify estimate shifts under varied covariate selections.
The Predictability-Computability-Stability (PCS) framework \citep{yu2020veridical,yu2024veridical,rewolinski2026pcs} synthesizes these concepts across the entire data science life cycle, formalizing stability as a core requirement for reliable inference.
As suggested by~\citet{short2026multicurious}, we will use the term multiverse analysis as an umbrella term for such types of stability analyses going forward.
Despite the appeal of mapping out many potential decisions, recent work suggests that evaluating a smaller set of theoretically grounded models often yields clearer insights than executing a massive multiverse \cite{delgiudice2021travelers, auspurg2025robustness}.
Nonetheless, the theoretical utility of the multiverse approach remains an active area of discussion \cite{rohrer2026whats}, supported by new preregistration protocols \cite{short2025systematic} and practitioner guides \cite{short2026multicurious}.
Most recently, this concept has expanded into automated data science, where AI agents are utilized to navigate multiverses in meta-analyses \cite{bertran2026aianalysts}.

\subsection{Understanding the Multiverse}

While multiverse analyses make analytical variability explicit, distilling their results into insights and communicating them remains an open problem with a large variety of approaches and lack of consensus.
By far the most common way to examine and report multiverse analyses is descriptively, typically in the form of histograms \citep[e.g.,][]{steegen2016increasing,simson2024one} and specification curves~\citep[e.g.,][]{simonsohn2020specification,godwin2025multiverse}.
Interactive visual systems such as Milliways \citep{sarma2024milliways} and BOBA \citep{liu2021boba} help researchers systematically evaluate complex data analysis paths.
However, as the decision space expands, these interactive tools fall short of quantitatively explaining the key drivers of variation within the multiverse.
To isolate the most important decisions, researchers often encode pipeline decisions as predictor variables~\citep[e.g.,][]{engzell2023understandinga,plessen2025digital,krahmer2026visualizing} and regress the output metrics onto them, using the resulting coefficients as proxies for decision influence \citep{young2025analytical}.
While computationally straightforward, this regression-based approach assumes that the decisions exert independent, additive effects on the outcome, and consequently fails to capture high-order interactions and non-linearities that are frequently present in real-world decision spaces.
These limitations can be addressed by applying more flexible models, e.g., a generalized functional ANOVA (fANOVA) \citep{hooker2007generalized} using random forests, which provides a nonparametric way to assess decision importances \citep{hutter2014efficient, simson2024one} or utilizing feature importance scores of a variety of other predictive models~\citep{short2026multicurious,bell2022modeling}.
However, this approach still relies on fitting a surrogate model to the multiverse and interpreting that model rather than analyzing the multiverse directly.
This abstraction introduces the risk of inaccurate importance scores if the surrogate model fails to capture the underlying signal of the multiverse.
Aside from model-based approaches, several works develop various one-off approaches to examining decision importance. Some examine the distribution of the metric of interest directly, e.g. by examining the absolute mean/min/max differences between different universes \citep{ferguson2025uncertaintya} or examining the fraction of universes where changing a single decision leads to an alternative outcome \citep{cassee2026exploringa}. Others use test statistics to estimate the effect of specific decisions, such as the F-test from a one-way fANOVA, the Kolmogorov-Smirnov statistic \citep{liu2021boba} or the K-samples Anderson–Darling test \citep{liu2023approximation}.


\subsection{Global Sensitivity Analysis}
Global sensitivity analysis (GSA) seeks to quantify how uncertainty in the output of a system can be attributed to uncertainty in its various inputs~\cite{saltelli2008global}.
In a multiverse, these inputs correspond to the set of analytical decisions.
Within GSA, variance-based Sobol' indices~\cite{sobol1993sensitivity,homma1996importance} use variance decomposition to break down output uncertainty into the sum of input effects and their interactions.
In contrast, \citet{borgonovo2007new} introduces a moment-independent measure which instead uses differences between distributions of outputs.
Similarly to how \method\ compares universes which differ in a single decision, \citet{morris1991factorial} proposes the elementary-effects method, which randomly selects points in the design space and averages one-at-a-time changes.
The closest related work is \citet{saltelli2025global}, which applies Sobol' indices to the many-analysts case study from \citet{breznau2022observing}, finding that most of the variation in the output is due to interactions between decisions.
Sobol' indices are a special case of the \method\ framework, specifically when it is instantiated with mean and variance on a full multiverse (Section~\ref{subsubsec:sobol-connection}).
\method's other summary and spread operators extend the same comparisons beyond importance in the average case, allowing for more flexible analyses such as worst-case vulnerability (Section~\ref{sec:beyond-average-case}).
We further provide explicit rules for handling sparse multiverses in tractable cases, such as merging functionally dependent decisions and restricting importance calculations to backgrounds where the decision of interest applies (Section~\ref{subsec:method-sparse}).
\section{Framework}
\label{sec:framework}

\begin{figure}[tb]
    \centering
    \begin{tikzpicture}[
        edge from parent/.style={draw, -latex, thin},
        level distance=0.85cm,
        level 1/.style={sibling distance=4.4cm},
        level 2/.style={sibling distance=2.2cm},
        level 3/.style={sibling distance=1.08cm},
        level 4/.style={sibling distance=1.08cm},
        ]
        \node[setupnode] {$\varnothing$}
        child { node[setupnode] {$c_1{=}1$}
          child { node[setupnodehigh] {$c_2{=}1$}
            child { node[setupnode] {$c_3{=}1$} child { node[setupleaf, lnode1] (lf1) {$f$} } }
            child { node[setupnode] {$c_3{=}2$} child { node[setupleaf, lnode2] (lf2) {$f$} } }
          }
          child { node[setupnodehigh] {$c_2{=}2$}
            child { node[setupnode] {$c_3{=}1$} child { node[setupleaf, lnode1] (lf3) {$f$} } }
            child { node[setupnode] {$c_3{=}2$} child { node[setupleaf, lnode2] (lf4) {$f$} } }
          }
        }
        child { node[setupnode] {$c_1{=}2$}
          child { node[setupnodehigh] {$c_2{=}1$}
            child { node[setupnode] {$c_3{=}1$} child { node[setupleaf, lnode3] (lf5) {$f$} } }
            child { node[setupnode] {$c_3{=}2$} child { node[setupleaf, lnode4] (lf6) {$f$} } }
          }
          child { node[setupnodehigh] {$c_2{=}2$}
            child { node[setupnode] {$c_3{=}1$} child { node[setupleaf, lnode3] (lf7) {$f$} } }
            child { node[setupnode] {$c_3{=}2$} child { node[setupleaf, lnode4] (lf8) {$f$} } }
          }
        };
        \draw[mvseta, line width=1.6pt, bend right=28] (lf1.south) to (lf3.south);
        \draw[mvsetb, line width=1.6pt, bend right=28] (lf2.south) to (lf4.south);
        \draw[mvsetc, line width=1.6pt, bend right=28] (lf5.south) to (lf7.south);
        \draw[mvsetd, line width=1.6pt, bend right=28] (lf6.south) to (lf8.south);
        \end{tikzpicture}
    \caption{A simple multiverse represented as a depth-3 tree, with each leaf node holding the result $f\left(\mathcal{U}\right)$ of one universe. Here, we can imagine $c_1$ being data cleaning (e.g., drop or impute missing values), $c_2$ being feature selection (e.g., include or exclude sensitive attributes), and $c_3$ being model choice (e.g., logistic regression or random forest). Taking $c_2$ as the \focustext{decision of interest}, each leaf node color marks one comparable set: the universes which agree on every decision but $c_2$. \method\ measures the spread $\Phi$ within each of these sets.}
    \label{fig:mv-tree}
\end{figure}
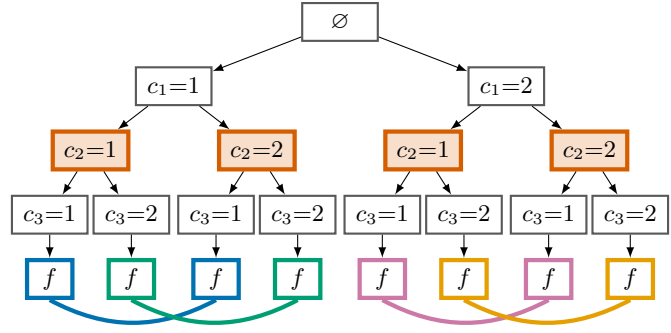

This section introduces the \method\ (\textbf{D}ecision \textbf{E}valuation by \textbf{C}alculating \textbf{I}mportance along \textbf{D}iverging \textbf{E}dges) framework, a flexible and model-free way to measure the importance of decisions in a multiverse and related analyses.
To properly understand how a multiverse is shaped by its decisions, we first formalize its geometry.
%
%
We then leverage the multiverse's geometry to detail the \method\ framework, including its intuition and theoretical connections.

\subsection{Multiverse Representations \& Notation}
\label{subsec:notation}

Here we provide a formal representation and notation for the multiverse.
To the best of our knowledge, this is the first time such a notation has been made available and we hope that it allows for a common language when reasoning about the multiverse and for methodological advancements in the analysis of data from multiverse analyses.

Consider a data science pipeline consisting of $d$ decisions.
Each decision $i\in\{1,\dots,d\}=[d]$ has a set of plausible options which we can enumerate as $\left\{1,\dots,o_i\right\}=[o_i]$.
We denote the option chosen for the $i$th decision as $c_i\in[o_i]$.
It follows that we can represent a universe $\mathcal{U}$ as a sequence of choices $(c_1,\dots,c_d)$, and the multiverse $\mathcal{M}$ as the Cartesian product $[o_1]\times\dots\times[o_d]$ of all possible universes, which spans $\mathbb{Z}_+^{o_1\times\dots\times o_d}$.
Every universe $\mathcal{U}\in\mathcal{M}$ is a distinct coordinate in this space, with some downstream result of interest denoted as $f(\mathcal{U})$.
For example, an analyst interested in model performance can define $f$ as accuracy, whereas one isolating a feature's causal effect can set $f$ to a specific regression coefficient.

We can map $\mathbb{Z}_+^{o_1\times\dots\times o_d}$, the space of valid universes, to the space of complete depth-$d$ decision trees; an example of this can be seen in Figure \ref{fig:mv-tree}.
In this characterization, the root node represents the pipeline before any decisions are made. It follows that the nodes at level $(i-1)$ will have $o_i$ children, resulting in $\prod_{i=1}^do_i$ leaf nodes, one for each $\mathcal{U}\in\mathcal{M}$.
To analyze the impact of the $i$th decision, we can decompose any universe $\mathcal{U}$ into two elements:
\begin{enumerate}
    \item The choice $c_i\in[o_i]$ made for the $i$th decision.
    \item The sequence of choices made for all other decisions, which we call the \textit{background} and denote as $b_{-i}\in\mathcal{B}_{-i}$, where $\mathcal{B}_{-i}=[o_1]\times\dots\times[o_{i-1}]\times[o_{i+1}]\times\dots\times[o_d]$.
\end{enumerate}
We can then express any universe as $\mathcal{U}=(c_i,b_{-i})$, isolating the $i$th decision from its background and allowing us to write the outcome as $f\left(c_i,b_{-i}\right)$.
We note that the numerical ordering $i\in[d]$ is purely for notational convenience; the \method\ framework does not require the decisions to have an inherent sequential ordering.
We also note that while we assume a full multiverse $\mathcal{M}$ of size $\prod_{i=1}^do_i$ here, in practice multiverses sometimes end up slightly smaller (i.e. \textit{sparse}) due to practical constraints, such as certain combinations of options being implausible or not technically feasible.
As outlined in Section~\ref{subsec:method-sparse}, the described method also applies for sparse multiverses.


\subsection{Generalized Spread \& Summary Functions}
\label{subsec:functions}

We now would like to use our geometric representation of a multiverse to quantify how much each decision drives its downstream variation.
Intuitively, switching the choice for an important decision should cause a change in the resulting universe's output.
To measure these changes across the multiverse, we will use two distinct functions: one which takes in a set of universes with identical backgrounds and measures the divergence in their results, and another which aggregates these variations across all backgrounds.
We define these operations using generalized spread and summary functions so that the framework can represent diverse applications and risk profiles.
\begin{itemize}
    \item \textit{The Spread Function $\Phi$.} Maps a set of outcomes to a value measuring their dispersion. While $\Phi$ can be variance, it can also describe spread in other ways, such as range, IQR, etc.
    \item \textit{The Summary Function $\Lambda$.} Maps a set of outcomes to a summary value. While $\Lambda$ can be the mean, it can also represent a quantile (e.g., median or maximum), the sum, etc.
\end{itemize}
%


Additionally, in many settings, not all decisions are equally plausible.
Analysts often have strong priors about which choices are most defensible, inducing a probability distribution over the multiverse $\mathcal{M}$.
Since $\Phi$ and $\Lambda$ are general operators, they can naturally incorporate these path-dependent probabilities.
Under standard moment-based definitions, this corresponds to $\Lambda$ as a weighted mean and $\Phi$ as a weighted variance.




\subsection{The \method\ Framework for Decision Importances}

We propose \method, a model-free decision importance framework which leverages multiverse structure to compare universes differing only in a particular decision.
Figure \ref{fig:mv-tree} illustrates the framework on a small multiverse, taking $c_2$ as the \focustext{decision of interest}.
The framework is defined in the following three steps:
\begin{enumerate}
    \item \textbf{Gather ``adjacent'' universes.} Universes $\mathcal{U}=(c_1,\dots,c_d)$ and $\mathcal{U}'=(c_1',\dots,c_d')$ are \textit{adjacent} at decision $i$ if $c_i\ne c_i'$ and $b_{-i}=b_{-i}'$. (In Figure \ref{fig:mv-tree}, universes adjacent at the \focustext{decision of interest} share a leaf node color.)
    \item \textbf{Measure the spread of ``comparable'' sets.} A set is \textit{comparable} if its member universes are pairwise-adjacent. For a given background $b_{-i}\in\mathcal{B}_{-i}$, this means isolating the variation caused by the $i$th decision using $\Phi\left(\left\{f\left(j,b_{-i}\right)\right\}_{j=1}^{o_i}\right)$. (Each color in Figure \ref{fig:mv-tree} marks one such comparable set.)
    \item \textbf{Aggregate across all spreads.} Applying $\Lambda$ over all background contexts $b_{-i}\in\mathcal{B}_{-i}$ yields: \[\method_{\texttt{total}}(i)=\Lambda_{b_{-i}\in\mathcal{B}_{-i}}\left(\Phi\left(\left\{f\left(j,b_{-i}\right)\right\}_{j=1}^{o_i}\right)\right).\]
\end{enumerate}

The \method\ framework applies the principle of comparison from causal inference, which tells us that if two units (universes) are alike except for the treatment (decision of interest), any difference in their outcomes can be attributed to the treatment.
Thus, by measuring the dispersion $\Phi$ of each comparable set, we get a full picture of the variation driven by each decision.
Furthermore, the \method\ framework has the following desirable properties.
\begin{itemize}[leftmargin=*]
    \item \textbf{\method\ is order-invariant.}
    Many analytical pipelines contain decisions which do not have a strict ordering.
    It follows that we do not want decision importances to change if the indices of the decisions are shuffled.
    Observe that because applicable measures of spread $\Phi$ (e.g. variance, range) and summary $\Lambda$ (e.g. sum, max) are invariant to the ordering of the input sets, we have that \method\ importances are invariant to decision ordering.
    
    \item \textbf{\method\ is risk-adaptable.}
    Different choices of summary function $\Lambda$ and spread function $\Phi$ allows \method\ to tailor importances to specific risk profiles.
    Practitioners can choose to evaluate expected behavior using traditional moments (mean and variance), isolate sensitivity to worst-case fluctuations (maximum and range), incorporate path-dependent prior probabilities (weighted mean and weighted variance), and more.
    We explore this risk adaptability further in Section~\ref{sec:beyond-average-case}, where we shift from measuring average-case influence to worst-case vulnerability.
    
    \item \textbf{\method\ is interaction-aware.}
    \method\ preserves the heterogeneous impacts of interactions by evaluating the spread induced by decisions locally (i.e. within each comparable set) before summarizing them globally (i.e. across all comparable sets).
    Interaction-specific importances can also be obtained by selecting a subset of decisions $\mathcal{S}\subseteq[d]$ and inspecting the universes that are adjacent at $\mathcal{S}$.
\end{itemize}

\subsection{Isolating Main and Interaction Effects with \method}
\label{subsec:method-main-interaction}
In complex multiverses, decisions may shift downstream results through a combination of fixed effects and background-dependent interactions.
To adequately capture these complexities, we first must extend the \method\ framework from analyzing single decisions to inspecting the importance of any subset $\mathcal{S}\subseteq[d]$.
Let $c_\mathcal{S}\in\prod_{i\in\mathcal{S}}[o_i]=\mathcal{C}_\mathcal{S}$ represent the choices for decisions in $\mathcal{S}$, and let $b_{-\mathcal{S}}\in\mathcal{B}_{-\mathcal{S}}$ represent the background decisions not in $\mathcal{S}$.
We then define the total importance of $\mathcal{S}$ as
\[
    \method_\texttt{total}(\mathcal{S}) =
    \Lambda_{b_{-\mathcal{S}} \in \mathcal{B}_{-\mathcal{S}}}
    \left(
        \Phi
        \left(
            \left\{
                f\left(
                    c_\mathcal{S},b_{-\mathcal{S}}
                \right)
            \right\}_{c_\mathcal{S}\in\mathcal{C}_\mathcal{S}}
        \right)
    \right).
\]
We refer to this quantity as the \textit{total effect} because it includes each of the ways decisions in $\mathcal{S}$ can induce variation in the downstream result:
\begin{itemize}
    \item \textit{Main effects.}
    Specific choices in $\mathcal{S}$ shift outcomes into different regimes, regardless of the background decisions.
    For example, consider an imputation method which consistently degrades downstream performance across all modeling options.
    
    \item \textit{Interaction effects.}
    Decisions in $\mathcal{S}$ only impact the downstream result in conjunction with choices made in the background.
    For example, consider a feature selection technique which only improves accuracy when used with a specific model class.
\end{itemize}
We can isolate the fixed effect of $\mathcal{S}$ by reversing the order with which we apply our generalized functions.
By collapsing the background variation before calculating the spread across different possible choices, we can calculate:
\[
    \method_\texttt{main}(\mathcal{S}) =
    \Phi_{c_\mathcal{S}\in\mathcal{C}_\mathcal{S}}
    \left(
        \Lambda
        \left(
            \left\{
                f(c_\mathcal{S}, b_{-\mathcal{S}})
            \right\}_{b_{-\mathcal{S}}\in\mathcal{B}_{-\mathcal{S}}}
        \right)
    \right).
\]
Changing the order of the functions $\Phi$ and $\Lambda$ establishes a conceptual distinction between the two metrics:
\begin{itemize}
    \item $\method_\texttt{total}(\mathcal{S})$ measures the spread which remains when everything except $\mathcal{S}$ is fixed.
    \item $\method_\texttt{main}(\mathcal{S})$ measures the spread caused by varying only $\mathcal{S}$, after summarizing across all backgrounds.
\end{itemize}

\subsection{Properties Under Mean-Variance Instantiations}
\label{subsec:properties}

While $\Lambda$ and $\Phi$ represent general summary and spread functions, the \method\ framework has special properties when instantiated with mean and variance, respectively.
Under this setting, the Law of Total Variance enables additive decompositions of the resulting decision importances.
We break down this particular case in two parts: first, by exploring the interpretations of total, main, and interaction effects, and second, by connecting \method\ to Sobol' indices, a well-studied form of sensitivity analysis.

\subsubsection{Interpretations Through Variance Decompositions}

Under the mean-variance characterization, the Law of Total Variance yields intuitive interpretations of $\method_\texttt{total}$ and $\method_\texttt{main}$.
We state these interpretations here and derive them in Section~\ref{subsec:supp-interp}.

\textbf{Total effect as remaining uncertainty.}
Through the Law of Total Variance, we have that
\begin{equation}
    \label{eq:total-effect-decomp}
    \frac{\method_\texttt{total}(\mathcal{S})}{\operatorname{Var}(\{f(\mathcal{U})\}_{\mathcal{U}\in\mathcal{M}})} =
    1 - \frac{\operatorname{Var}\left(\mathbb{E}\left[f(\mathcal{U})\mid b_{-\mathcal{S}}\right]\right)}{\operatorname{Var}(\{f(\mathcal{U})\}_{\mathcal{U}\in\mathcal{M}})}.
\end{equation}
The left-hand side of Equation~\ref{eq:total-effect-decomp} quantifies the \textit{proportion of total variance remaining} after fixing every decision except those in $\mathcal{S}$.
Thus, $\method_\texttt{total}(\mathcal{S})$ measures the entirety of the variance controlled by $\mathcal{S}$, including both its main effect and its interactions with other decisions.

\textbf{Main effect as variance elimination.}
Again applying the Law of Total Variance, we have that
\begin{equation}
    \label{eq:main-effect-decomp}
    \frac{\method_\texttt{main}(\mathcal{S})}{\operatorname{Var}(\{f(\mathcal{U})\}_{\mathcal{U}\in\mathcal{M}})} =
    1 - \frac{\mathbb{E}\left[\operatorname{Var}(f(\mathcal{U})\mid c_\mathcal{S})\right]}{\operatorname{Var}(\{f(\mathcal{U})\}_{\mathcal{U}\in\mathcal{M}})}.
\end{equation}
Observe that the right-hand side of Equation~\ref{eq:main-effect-decomp} measures the proportion of variance left over after making choices for the decisions in $\mathcal{S}$.
It follows that the left-hand side of Equation~\ref{eq:main-effect-decomp} represents the \textit{proportion of total variance eliminated by fixing the decisions in $\mathcal{S}$}.
This tells us that a high main effect (relative to total variance) indicates that choices $c_\mathcal{S}$ reliably shift outcomes into distinct regimes, regardless of the background choices in the pipeline.

\textbf{The main-to-total ratio.}
Since $\method_{\texttt{total}}(S)$ encompasses the fixed effect of $S$ and every interaction it participates in, we can define the interaction effect of $S$ as:
\[
    \method_\texttt{interactions}(\mathcal{S}) = \method_\texttt{total}(\mathcal{S}) - \method_\texttt{main}(\mathcal{S}).
\]
In Appendix~\ref{subsec:supp-interp} we show that this residual is non-negative and consists exactly of the interactions involving $S$.
It follows that evaluating the ratio of main effect to total effect gives us an intuitive measure for $\mathcal{S}$'s influence:
\begin{align*}
    \operatorname{Ratio}_\texttt{main}(\mathcal{S}) &=
    \frac{\method_\texttt{main}(\mathcal{S})}{\method_\texttt{total}(\mathcal{S})}\\
    &= \frac{\method_\texttt{main}(\mathcal{S})}{\method_\texttt{main}(\mathcal{S}) + \method_\texttt{interactions}(\mathcal{S})}.
\end{align*}
We thus have that $\operatorname{Ratio}_\texttt{main}(\mathcal{S})$ characterizes how decisions in $\mathcal{S}$ influence the multiverse.
If $\operatorname{Ratio}_\texttt{main}(\mathcal{S})$ is close to one, then the decisions in $\mathcal{S}$ operate independently and tend to shift performance uniformly, but if it is close to zero, then the impact of $\mathcal{S}$ depends more on interactions with background decisions.

\subsubsection{Theoretical Connections to Sobol' Indices}
\label{subsubsec:sobol-connection}

Staying in the special case where $\Lambda$ represents the mean and $\Phi$ represents the variance, the normalized importance scores output by the \method\ framework map cleanly onto Sobol' indices, a measure of variance-based sensitivity.
This equivalence, which we derive in Appendix~\ref{subsec:supp-sobol}, allows \method\ to inherit theoretical properties from the variance-based sensitivity analysis.
Specifically:
\begin{itemize}
    \item $\method_\texttt{main}(\mathcal{S})/\operatorname{Var}(\{f(\mathcal{U})\}_{\mathcal{U}\in\mathcal{M}})$ corresponds to the first-order Sobol' index for $|\mathcal{S}|=1$ and the closed Sobol' index for $|\mathcal{S}|\geq2$.
    The first-order Sobol' index measures the proportion of the total downstream variance caused by a single variable acting on its own.
    The closed Sobol' index additionally captures the interactions among the decisions of $S$.
    
    \item $\method_\texttt{total}(\mathcal{S})/\operatorname{Var}(\{f(\mathcal{U})\}_{\mathcal{U}\in\mathcal{M}})$ corresponds to the total-effect Sobol' index.
    This index measures the full contribution of a variable to the downstream variance, including its participation in interactions.
\end{itemize}
Classical Sobol' indices are limited to moment-based statistics (i.e., mean and variance), which may limit their applicability when the outcome distribution is skewed or the setting is risk-sensitive.
While the exact additivity and formal equivalence to Sobol' indices hold only under mean and variance, \method\ generalizes this core sensitivity analysis measure to arbitrary summary and spread operators.

\subsection{Extending \method\ to Sparse Multiverses}
\label{subsec:method-sparse}

In Section~\ref{subsec:notation}, we assume the multiverse is \textit{full}, meaning that $\mathcal{M}$ is the Cartesian product $[o_1]\times\dots\times[o_d]$.
That is, each decision $i\in[d]$ has the same $o_i$ options available in every branch of the multiverse, regardless of the choices made elsewhere.
However, in practice, multiverses are often \textit{sparse}: some combinations of choices are implausible, and some decisions or options only exist in certain branches of the multiverse.
Formally, $\mathcal{M}$ is then a proper subset of $[o_1]\times\dots\times[o_d]$.
Recall that \method\ relies on `comparable sets' of universes that share a background $b_{-i}$ and differ only in decision $i$.
In a sparse multiverse, these sets may be incomplete, and how the gaps are handled can affect the resulting importances.
Here, we describe three types of sparsity and how each is treated.
Figure~\ref{fig:sparse-types} in Appendix~\ref{sec:supp-figures} displays visualizations corresponding to each sparsity type.

\textbf{Functional dependence} (Figure~\ref{subfig:sparsity-functional-dependence})\textbf{.}
This occurs when the choice for one decision fully determines the choice for another, e.g., if the data cleaning procedure fixes the feature set used by a model.
Since these decisions cannot be varied separately, their impacts cannot be distinguished.
We therefore merge them and report a single importance.

\textbf{Partially-applicable decisions} (Figure~\ref{subfig:sparsity-partially-applicable})\textbf{.}
Some decisions only apply under certain background choices.
For example, the choice between Bayesian and frequentist methods determines whether the choice of prior is relevant.
Including frequentist backgrounds would bias the prior's importance toward zero, since priors do not affect frequentist models.
We therefore compute a decision's importance only over the backgrounds where it applies, and additionally report the proportion of the multiverse this covers.

\textbf{Background-dependent options} (Figure~\ref{subfig:sparsity-background-dependent})\textbf{.}
Some options may only be valid in certain branches of the multiverse.
For example, consider a dataset consisting of both adults and minors, where restricting the sample to minors allows for an additional feature set which includes `parental income'.
When computing the importance of the sample restriction, we are unable to compare a universe including `parental income' to its counterfactual, and thus do not have a complete comparable set.
Therefore, we impute the missing universes with their best-matched counterfactual.
Specifically, we take the summary $\Lambda$ over the full-sample (i.e., non-minor-restricted) universes that agree with every decision except the feature set, which cannot be matched.
%
%
Due to this imputation, the variance decomposition from Section~\ref{subsec:properties} holds only approximately in this case.
\section{Issues with Model-Dependent Decision Importances}
\label{sec:model-dependent}

\begin{figure*}[tb]
    \centering
    \subcaptionbox{\label{fig:fertility-obs-pred-hist}}
        [0.62\linewidth]{\includegraphics[width=\linewidth]{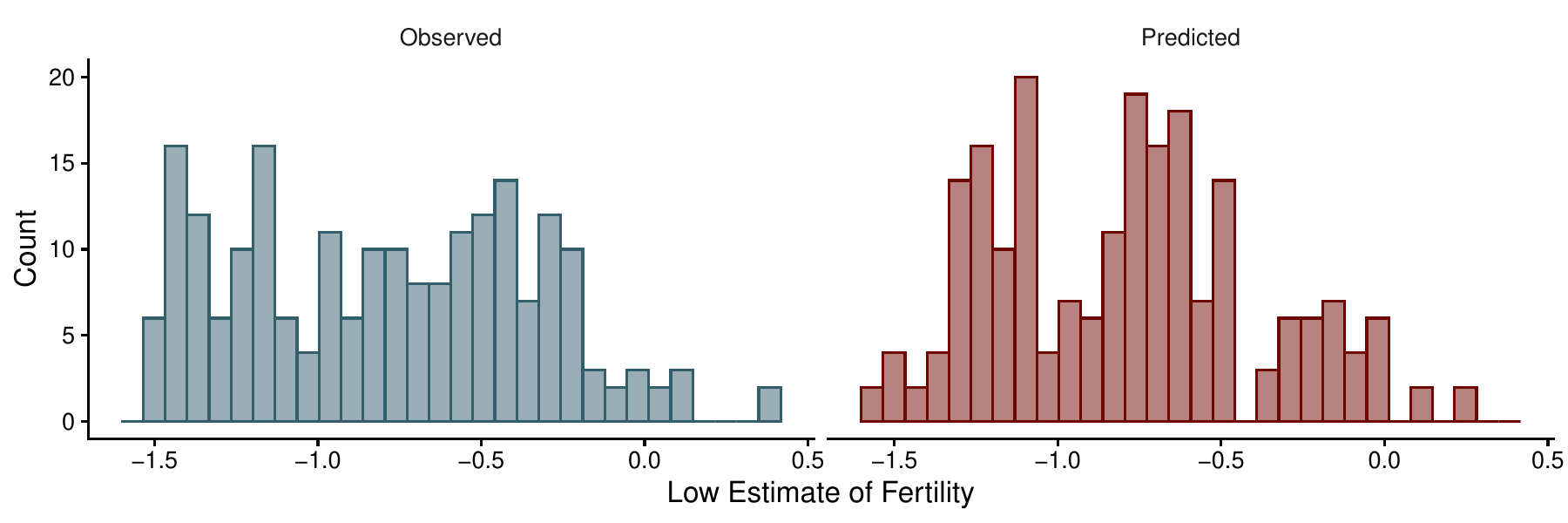}}
    \subcaptionbox{\label{fig:fertility-resid-fitted}}
        [0.3\linewidth]{\includegraphics[width=\linewidth]{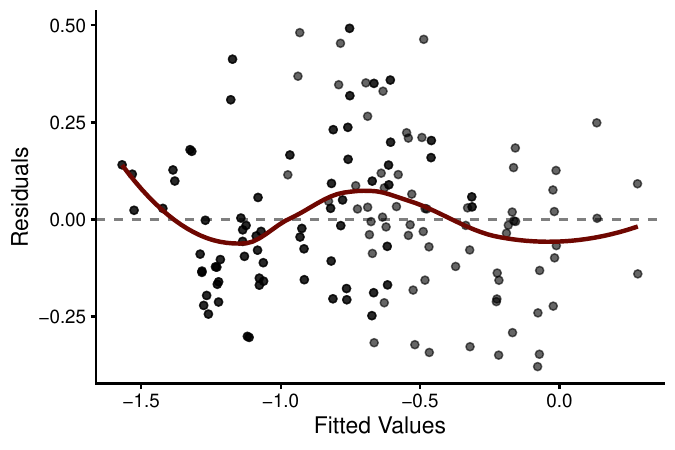}}
        
    \subcaptionbox{\label{fig:hurricane-obs-pred-hist}}
        [0.62\linewidth]{\includegraphics[width=\linewidth]{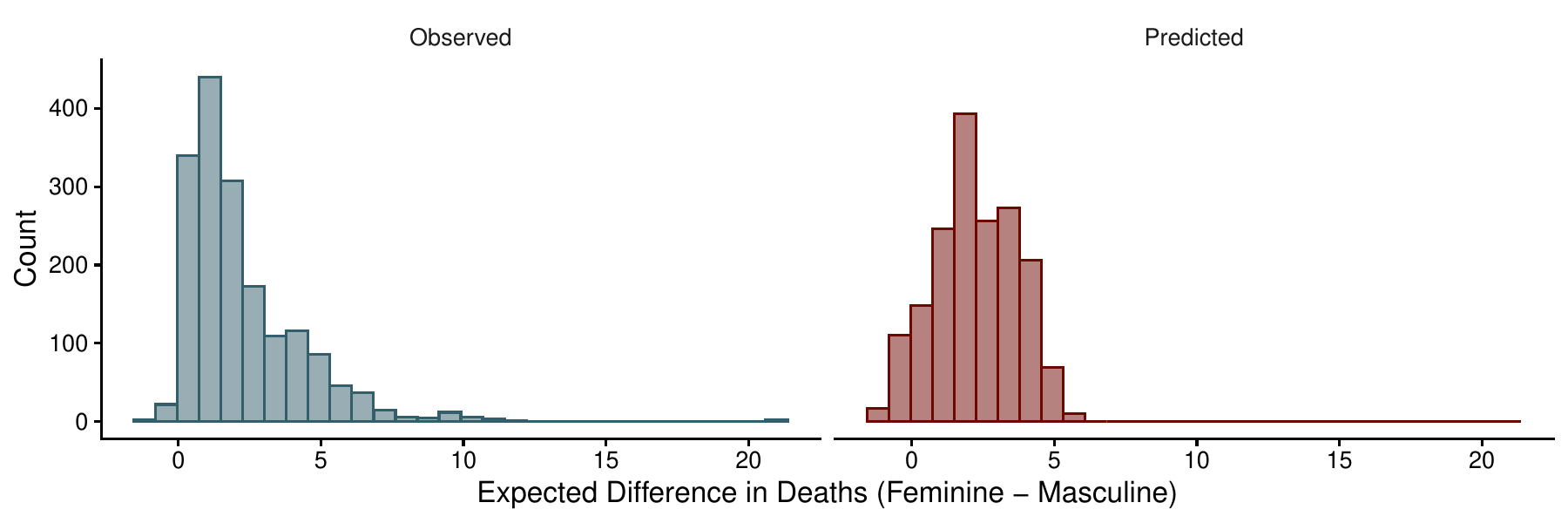}}
    \subcaptionbox{\label{fig:hurricane-resid-fitted}}
        [0.3\linewidth]{\includegraphics[width=\linewidth]{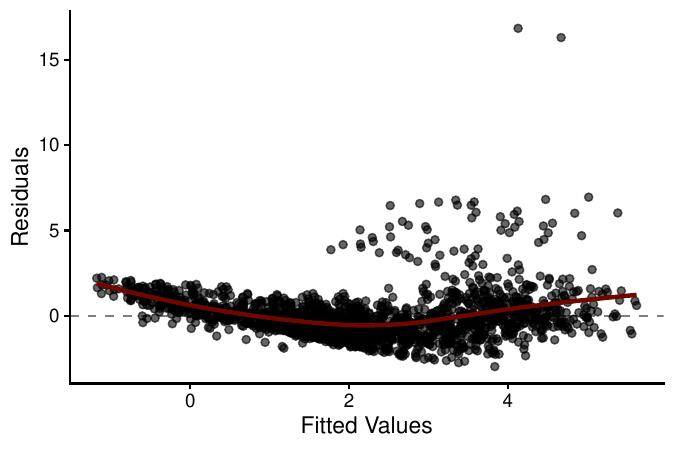}}
    \caption{Surrogate regression models poorly approximate the multiverse outcomes they are used to explain. \textbf{Left:} observed vs. predicted distributions of universe outcomes. \textbf{Right:} residuals vs. fitted values. \textbf{Top:} on \textit{fertility}, the predicted distribution has a different shape than the observed outcomes. \textbf{Bottom:} on \textit{hurricane}, the surrogate misses the long right tail of observed outcomes and the residuals show heteroscedasticity, outliers, and curvature.}
    \label{fig:regression-diagnostics}
\end{figure*}

A primary advantage of the \method\ framework is its ability to avoid the pitfalls of proxy-based alternatives such as regression or fANOVA.
These approaches rely on fitting a surrogate model $\hat{f}$ to predict the observed outcome $f(\mathcal{U})$ of some universe $\mathcal{U}$.
However, assuming a fitted predictive model and its interpretations accurately represent a deterministic multiverse introduces weaknesses.
To show this, we begin by discussing the~underlying assumptions behind the use of surrogate models and why they are a poor choice for analyzing multiverse decisions.
We then empirically demonstrate that surrogate models fail to properly fit real-world multiverse data.
Finally, we use a simulation to show that model-based decision importances systematically misattribute influence unless forced to perfectly extrapolate the data, at which point they collapse into computationally expensive approximations of \method.

\subsection{Why Choose to Interpret a Surrogate?}

In standard supervised learning settings, predictive models are fit as proxies for reality out of necessity.
Practitioners have a finite sample of training data and are interested in unseen test points whose ground-truth response is unknown.
In such settings, fitting a model is a practical way to bridge this gap and make inferences about these unknown points.

However, this is not the setting we find ourselves in when conducting a multiverse analysis.
By definition, a multiverse explicitly enumerates all plausible analysis pipelines, so there are no unobserved universes of interest.
Fitting a model to this complete population only introduces unnecessary model-specific biases and approximation error.

While surrogate models are successfully used in fields like physics (where they are referred to as \textit{emulators}) to predict experimental outcomes across parameter grids, their success relies on underlying phenomena that follow well-understood physical laws.
This is notably not the case in applied data analysis, where the impact of analyst decisions on downstream results often exhibits highly non-linear and unpredictable relationships that fitted models can struggle to capture.
Instead of relying on a potentially flawed model to learn the structure of the multiverse, \method\ directly evaluates the impacts of each decision from the known ground-truth data.

\subsection{Evaluating Model Fit}
\label{subsec:eval-model-fit}

In traditional supervised learning, analysts are encouraged to inspect the ``goodness-of-fit'' of their model before it is used for inference.
Yu and Kumbier \cite{yu2020veridical} advocate for using predictability as a reality-check, ensuring that models are only used to draw conclusions if they reflect reality.
This is a common-sense procedure: if the model does not accurately capture the signal in the data, its explanations and insights are likely shaky \cite{freedman1991statistical,breiman2001statistical}.
We follow this logic to assess whether the models used to interpret multiverses (e.g., regression) produce trustworthy representations.

\begin{figure}[t]
    \centering
    \includegraphics[width=0.78\linewidth]{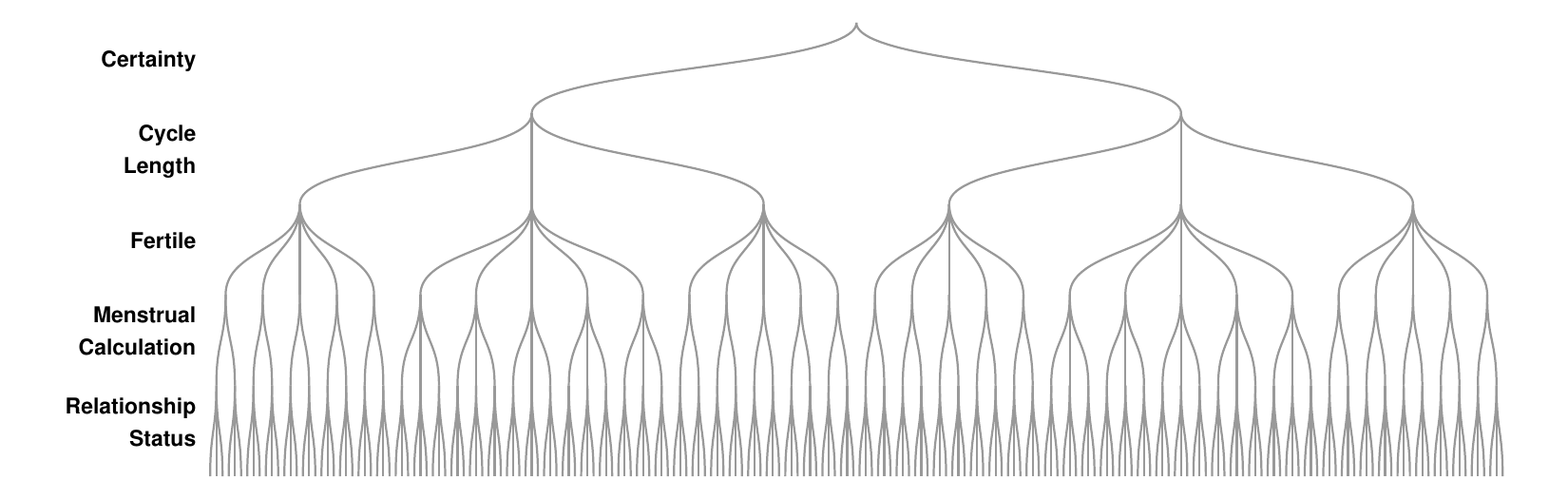}
    \caption{Full tree view of the \textit{Fertility / Durante} multiverse. Its branch structure expands over its five decisions, making explicit the full set of possible universe paths.}
    \label{fig:fertility-multiverse-tree}
\end{figure}

We evaluate the empirical fit of linear regression models on two well-studied multiverses from the literature: The \textit{fertility} multiverse (visualized in Figure~\ref{fig:fertility-multiverse-tree}) from Steegen et al. \citep{steegen2016increasing} based on the work of Durante et al. \citep{durante2013fluctuating}, which looks into the relationship between fertility and voting behavior, and the \textit{hurricane} multiverse, which focuses on whether hurricanes with feminine names lead to fewer precautionary measures \citep{jung2014female,malter2014female,bakkensen2014population,simonsohn2020specification}.
We describe these multiverses in greater detail in Section~\ref{subsec:fertility-durante} and Section~\ref{subsec:hurricane}, respectively.
By comparing the distributions of the observed universe outcomes to the model's predictions, we can test whether the proxy model is accurately capturing reality.

The diagnostic results shown in Figure~\ref{fig:regression-diagnostics} demonstrate that the regression models serve as poor proxies for the actual multiverse data.
The predictive accuracy of these models is highly variable: the adjusted $R^2$ between $f$ and $\hat{f}$ is 0.816 for the \textit{fertility} multiverse and 0.490 for the \textit{hurricane} multiverse.
Furthermore, the residual plots display non-random patterns that indicate the assumptions underlying valid linear regression models are violated.
This implies that heavy caution should be exercised when using model-based inferences.

\subsection{Failure Modes of Proxy Models}

\begin{figure}[tb]
    \centering
    \begin{tikzpicture}[
        edge from parent/.style={draw, -latex, thin},
        level distance=0.85cm,
        level 1/.style={sibling distance=4.4cm},
        level 2/.style={sibling distance=2.2cm},
        level 3/.style={sibling distance=1.08cm},
        level 4/.style={sibling distance=1.08cm},
        ]
        \node[setupnode] {$\varnothing$}
        child { node[setupnode] {$c_1{=}1$}
          child { node[setupnode] {$c_2{=}1$}
            child { node[setupnode] {$c_3{=}1$} child { node {0.825} } }
            child { node[setupnode] {$c_3{=}2$} child { node {0.865} } }
          }
          child { node[setupnode] {$c_2{=}2$}
            child { node[setupnode] {$c_3{=}1$} child { node {0.775} } }
            child { node[setupnode] {$c_3{=}2$} child { node {0.815} } }
          }
        }
        child { node[setupnode] {$c_1{=}2$}
          child { node[setupnode] {$c_2{=}1$}
            child { node[setupnode] {$c_3{=}1$} child { node {0.775} } }
            child { node[setupnode] {$c_3{=}2$} child { node {0.815} } }
          }
          child { node[setupnode] {$c_2{=}2$}
            child { node[setupnode] {$c_3{=}1$} child { node {0.825} } }
            child { node[setupnode] {$c_3{=}2$} child { node {0.865} } }
          }
        };
    \end{tikzpicture}
    \caption{A multiverse consisting of three binary decisions. The first two levels ($c_1$ and $c_2$) impact the output accuracies through their interaction. The last level ($c_3$) influences the accuracies with its fixed effect.}
    \label{fig:mv-proxy-counterexample}
\end{figure}
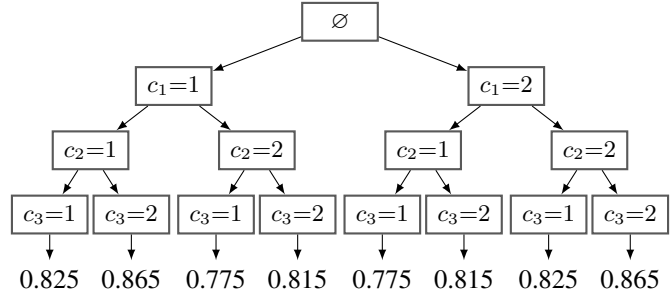

Since the ground-truth decision importances of real-world multiverses are unknown, we construct a simulated multiverse to demonstrate how proxy models can produce incorrect importance scores.
Consider the minimal toy multiverse depicted in Figure~\ref{fig:mv-proxy-counterexample}, which consists of three binary decisions.
The first two decisions act only through their pairwise interaction, while the third decision acts purely through its main effect.
No matter the background context, switching $c_1$ or $c_2$ shifts the outcome accuracy by 5\%, whereas switching $c_3$ shifts the accuracy by 4\%.
Thus, we have that the true importance of $c_1$ and $c_2$ is higher than that of $c_3$.

To test how various surrogates perform against \method, we compare against three proxy approaches.
First, we fit a linear regression model without interaction terms.
Omitting interaction terms is a fair and standard baseline because expecting an analyst to know the relevant interactions \textit{a priori} is unrealistic for large multiverses.
While an analyst could attempt to get around this by including all possible interaction terms, doing so saturates the regression model (the number of features $p$ becomes equal to the number of data points $n$), leaving zero residual degrees of freedom and making inference impossible.
Second, we fit a random forest and evaluate its feature importances according to both mean decrease in impurity (MDI) and permutation feature importance (PFI).
The forest is configured using the defaults from the popular Python package \texttt{scikit-learn}.
Finally, we include functional ANOVA (fANOVA) from Hutter et al. \citep{hutter2014efficient}, which was used by Simson et al. \citep{simson2024one} to infer decision importances.
This approach computes both main and interaction importances over the predictions of a similarly configured random forest (using the default implementation from Hutter et al.'s \texttt{fanova} package \citep{hutter2014efficient}).
Similarly to \method, we report fANOVA scores for both main effects (i.e., the importance of a decision in isolation) and total importance (i.e., including interaction terms involving the decision).
Since MDI, PFI, and fANOVA depend on the forest's randomness, we average their scores over ten seeds.

\begin{table}[t]
    \centering
    \caption{Comparison of decision importance scores on the multiverse shown in Figure~\ref{fig:mv-proxy-counterexample}. Only $\method_\texttt{total}$ correctly identifies that the interacting decisions ($c_1, c_2$) have a higher importance than the main effect decision ($c_3$). The most important feature(s) for each approach are shown in bold.}
    \label{tab:toy-importance-comparison}
\begin{tabular}{lcccccc}
\toprule
 & \multicolumn{2}{c}{} & \multicolumn{2}{c}{fANOVA} & \multicolumn{2}{c}{Random Forest} \\
\cmidrule(lr){4-5} \cmidrule(lr){6-7}
 & $\method_{\texttt{total}}$ & Reg. & Main & Total & MDI & PFI \\
\midrule
$c_1$ & \textbf{0.000625} & 0.000000 & 0.106792 & 0.355601 & 0.294151 & 0.000544 \\
$c_2$ & \textbf{0.000625} & 0.000000 & 0.116994 & 0.354927 & 0.283771 & 0.000534 \\
$c_3$ & 0.000400 & \textbf{0.040000} & \textbf{0.479171} & \textbf{0.627509} & \textbf{0.422078} & \textbf{0.000727} \\
\bottomrule
\end{tabular}
\end{table}

Table~\ref{tab:toy-importance-comparison} displays the results of each decision importance measure on the simulated multiverse from Figure~\ref{fig:mv-proxy-counterexample}.
We observe that $\method_\texttt{total}$ is the only decision importance metric which correctly ranks the influence of the three decisions, with each of the proxy models stating that $c_3$ is the most influential choice on the resulting accuracy.
These failures stem from the inherent mechanics of each proxy model:
\begin{itemize}
    \item Linear regression and $\text{fANOVA}_{\texttt{main}}$ fail because the pure interaction between $c_1$ and $c_2$ yields no individual effects. Thus, the regression model fails to assign them any importance, and $\text{fANOVA}_{\texttt{main}}$ assigns them a small amount that is only nonzero because of the imperfect fit of the random forest.
    \item MDI scores on the random forest rely on greedy splits. Since splitting on a feature which only impacts the result through an interaction will not reduce impurity at the root node, the main-effect decision $c_3$ dominates the initial splits and thus receives a disproportionate amount of credit.
    \item PFI and $\text{fANOVA}_{\texttt{total}}$ fail due to bootstrapping. Since some universes are out-of-bag for each tree, the ensemble must predict their outcomes without having seen them during training. As a result, these trees fall back to the global mean. This shrinkage to the mean causes both PFI and fANOVA to underestimate the importance of the interaction. Moreover, because fANOVA evaluates marginal variance on the surrogate model's predictions $\hat{y}$ rather than the observed outcomes $y$, its scores are tied to the errors of the forest.
\end{itemize}
The only way that proxy models can capture the correct importances in these interaction-heavy scenarios is to fit a random forest where each tree uses the full data and considers all features at every split.
Under this setup, the random forest perfectly interpolates the multiverse, allowing PFI and $\text{fANOVA}_{\texttt{total}}$ to recover the correct decision importance rankings.
However, under these interpolating conditions, PFI and $\text{fANOVA}_{\texttt{total}}$ are not distinct from our model-free approach---they collapse directly into \method, as we show in Appendix~\ref{subsec:supp-pfi-fanova}.
Thus, we have that interpreting proxy models is a lose-lose proposition: they either fail to faithfully capture the multiverse's signal, or they require computationally expensive proxy models to match the exact solution that \method\ directly provides.
\section{Real-Data Case Studies}
\label{sec:ablation}


In this section we apply regression coefficients, fANOVA, and \method\ to five different multiverse analyses and compare the results.
The results in this section use \method\ instantiated with $\Lambda=$ mean and $\Phi=$ variance.

Evaluating decision importance methods is challenging due to the lack of ground-truth importances in real-world data analysis pipelines.
To overcome this, we assess the stability of regression coefficients, fANOVA, and \method\ when a decision is removed from the pipeline.
Where the method supports it, we compare both \textit{total} (incl. interactions) and \textit{main} (excl. interactions) importance estimates.
Rather than debating if one set of rankings is more valid than another, this ablation provides an external criterion: approaches that better capture the signal of the full multiverse should provide more consistent importance profiles after these reductions.

\subsection{Ablation Setup}
\label{subsec:ablation-setup}

We adopt the notation from Section~\ref{sec:framework}.
Let the full multiverse be $\mathcal{M} = [o_1] \times \dots \times [o_d]$,
where decision $i \in [d]$ has options $[o_i]$, and each universe $\mathcal{U}$ is represented by a sequence $(c_1,\dots,c_d)$ with $c_i \in [o_i]$.
For each importance estimator $m$, we compute importances on the full multiverse
\[
\mathbf{I}^{\mathrm{full}}_m = \left(I^{\mathrm{full}}_{m,1}, \dots, I^{\mathrm{full}}_{m,d}\right),
\]
where $I^{\mathrm{full}}_{m,i}$ denotes the estimated importance of decision $i$ under method $m$.
We then generate ablated multiverses by fixing decision $i$ to one specific option $k \in [o_i]$, yielding the restricted multiverse
\[
\mathcal{M}_{i \leftarrow k} = \left\{\mathcal{U} = (c_1,\dots,c_d) \in \mathcal{M} : c_i = k\right\}.
\]
Within $\mathcal{M}_{i \leftarrow k}$, the $i$th decision no longer varies and therefore no longer receives an estimated importance score.
Re-estimating importances on the reduced multiverse yields
\[
\mathbf{I}^{(i \leftarrow k)}_m = \left(I^{(i \leftarrow k)}_{m,j} : j \in [d]\setminus\{i\}\right).
\]
To compare this ablated vector to the full-multiverse estimate, we also define the full vector with the $i$th entry removed:
\[
\mathbf{I}^{\mathrm{full}}_{m,-i} = \left(I^{\mathrm{full}}_{m,j} : j \in [d]\setminus\{i\}\right).
\]

The ablation analysis compares methods in two complementary ways: agreement across different methods (inter-method stability) and stability within a method (intra-method stability).

\subsubsection{Inter-Method Stability}
\label{subsubsec:inter-method-stability}

The first comparison asks whether different importance estimators induce similar rankings of the remaining decisions, both for the full multiverse and after a given ablation.
For the full multiverse $ \mathcal{M} $ and each ablated multiverse $\mathcal{M}_{i \leftarrow k}$, we compare the importance vectors produced by each pair of methods $(m_x,m_y)$ using the Spearman rank correlation
\[
\rho_{m_x,m_y}^{(i \leftarrow k)} =
\mathrm{Cor}_{\mathrm{Spearman}}\!\left(
\mathbf{I}^{(i \leftarrow k)}_{m_x},
\mathbf{I}^{(i \leftarrow k)}_{m_y}
\right).
\]
We report these choice-specific quantities averaged over all $k \in [c_i]$ for a fixed decision $i$.

\subsubsection{Intra-Method Stability}
\label{subsubsec:intra-method-stability}

The second comparison evaluates each method's internal robustness.
Rather than comparing methods, we compare a method's full-multiverse ranking to its post-ablation ranking.
For each method $m$ and ablation $(i \leftarrow k)$, we compute
\[
\rho_{m}^{(i \leftarrow k)} =
\mathrm{Cor}_{\mathrm{Spearman}}\!\left(
\mathbf{I}^{\mathrm{full}}_{m,-i},
\mathbf{I}^{(i \leftarrow k)}_m
\right).
\]
High intra-method stability indicates that an estimator preserves a similar ordering of the remaining decisions even when one source of multiverse variation is held fixed, while low stability indicates that the estimated importances depend strongly on the inclusion of other sources of variation.

\subsection{Fertility / Durante}
\label{subsec:fertility-durante}

\begin{table}[t]
\setlength{\tabcolsep}{4.5pt}
\caption{\label{tab:importance-fertility}Importance estimates per decision, by method for the \textit{Fertility / Durante} multiverse \citep{durante2013fluctuating,steegen2016increasing}.}
\centering
\begin{tabular}[t]{lrrrrrr}
\toprule
\multicolumn{1}{c}{ } & \multicolumn{3}{c}{\textsc{Decide}} & \multicolumn{2}{c}{fANOVA} & \multicolumn{1}{c}{ } \\
\cmidrule(l{3pt}r{3pt}){2-4} \cmidrule(l{3pt}r{3pt}){5-6}
Decision & Total & Main & Ratio & Total & Main & Reg.\\
\midrule
Relationship Status & 0.089 & 0.078 & 88.3\% & 0.506 & 0.410 & 0.297\\
Fertile & 0.087 & 0.056 & 64.0\% & 0.417 & 0.251 & 0.244\\
Menstrual Calculation & 0.041 & 0.011 & 27.1\% & 0.193 & 0.055 & 0.152\\
Cycle Length & 0.028 & 0.012 & 43.7\% & 0.132 & 0.050 & 0.152\\
Certainty & 0.008 & 0.005 & 69.6\% & 0.054 & 0.022 & 0.146\\
\bottomrule
\end{tabular}
\end{table}

In the paper introducing multiverse analyses, Steegen et al. \citep{steegen2016increasing} demonstrate the method using a multiverse analyzing the relationship between fertility (human reproductive cycle) and voting behavior, building on an original study by Durante et al. \citep{durante2013fluctuating}. Commonly referred to under the names \textit{Fertility} or \textit{Durante}, this multiverse consists of five decisions displayed in Table~\ref{tab:fertility-mv-info} of Appendix~\ref{sec:supp-setup}.
Figure~\ref{fig:fertility-ablation-inter-intra} visualizes the two ablation comparisons introduced above.

\begin{figure}[t]
    \centering
    \subcaptionbox{Inter-method stability: Spearman correlation between pairs of methods on the full multiverse (top-right) and averaged over all ablations (bottom-left).\label{fig:fertility-ablation-inter-method}}
        [3.84in]{\includegraphics[width=\linewidth]{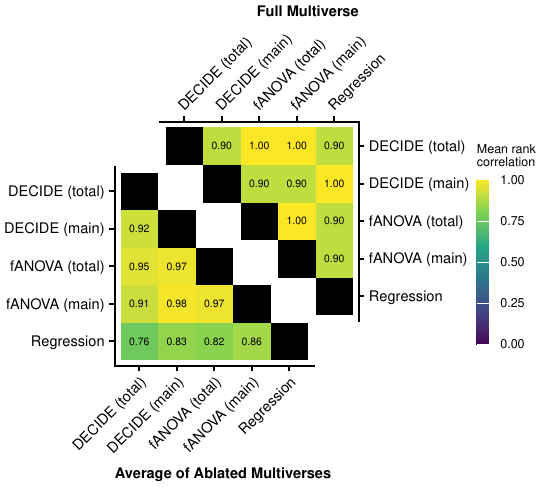}}
    \hfill
    \subcaptionbox{Intra-method stability: mean Spearman correlation between each method's full-multiverse and ablated rankings.\label{fig:fertility-ablation-intra-method}}
        [2.56in]{\includegraphics[width=\linewidth]{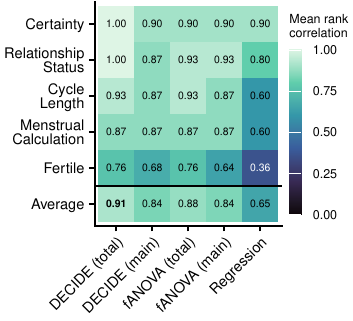}\vspace{0.2in}}
    \caption{Ablation stability on the \textit{Fertility / Durante} multiverse.
    \textbf{Left:} all pairs agree closely on the full multiverse ($\rho \ge 0.90$); after ablation, disagreement is concentrated in pairs including regression as an estimator.
    \textbf{Right:} $\method_{\texttt{total}}$ is the most stable estimator overall and most stable or tied for most stable in every ablation, regression is the least.}
    \label{fig:fertility-ablation-inter-intra}
\end{figure}

Table~\ref{tab:importance-fertility} reports the importance estimates of all three methods on the full multiverse.
All three agree on the overall ordering, placing `Relationship Status' first and `Fertile' second, but disagree by how clearly decisions are separated: the regression compresses the remaining three into a near-tie ($0.152$, $0.152$, $0.146$), whereas \method\ and fANOVA both clearly disambiguate them.
The \method\ `Ratio' column reports each decision's main effect as a share of its total effect.
This shows \textit{how} a decision acts; for example, `Relationship Status' is mostly a main effect ($88.3\%$), while `Menstrual Calculation' ($27.1\%$) and `Cycle Length' ($43.7\%$) influence the outcome mostly through interactions.

Averaged over the five ablations, $\method_{\texttt{total}}$ is the most stable estimator ($\bar{\rho} = 0.91$), followed by $\text{fANOVA}_{\texttt{total}}$ ($0.88$), $\method_{\texttt{main}}$ and $\text{fANOVA}_{\texttt{main}}$ (both $0.84$), while regression-based importances are considerably less stable ($0.65$).
Fixing `Fertile', the decision with the largest number of choices, is the hardest ablation for every estimator, yet degradation is far more severe for regression ($0.36$) than for $\method_{\texttt{total}}$ ($0.76$).
Inter-method agreement is structured accordingly: on the full multiverse all pairs agree closely ($\rho \ge 0.90$), and after ablation the four cross-method pairs formed by \method\ and fANOVA continue to agree closely with one another ($\bar{\rho}$ between $0.91$ and $0.98$), whereas every pair involving regression falls between $0.76$ and $0.86$, with regression and $\method_{\texttt{total}}$ disagreeing most.
Overall, $\method_{\texttt{total}}$ yields the highest intra-method stability, demonstrating that its importances are robust to the inclusion/exclusion of particular decisions.

\subsection{Hurricane / Jung}
\label{subsec:hurricane}

\begin{figure}[tb]
    \centering
    \subcaptionbox{Inter-method stability: Spearman correlation between pairs of methods on the full multiverse (top-right) and averaged over all ablations (bottom-left).\label{fig:hurricane-ablation-inter-method}}
        [3.84in]{\includegraphics[width=\linewidth]{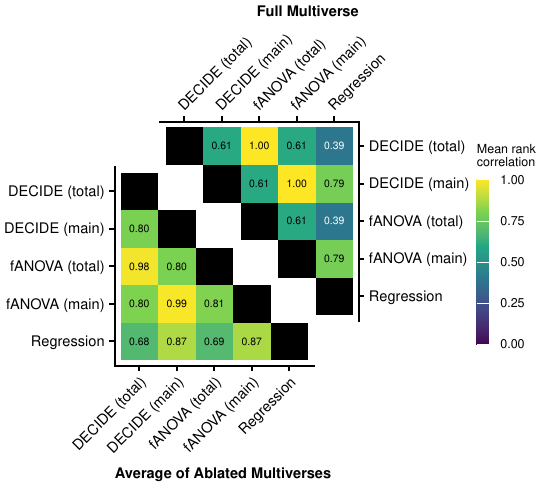}}
    \hfill
    \subcaptionbox{Intra-method stability: mean Spearman correlation between each method's full-multiverse and ablated rankings.\label{fig:hurricane-ablation-intra-method}}
        [2.56in]{\includegraphics[width=\linewidth]{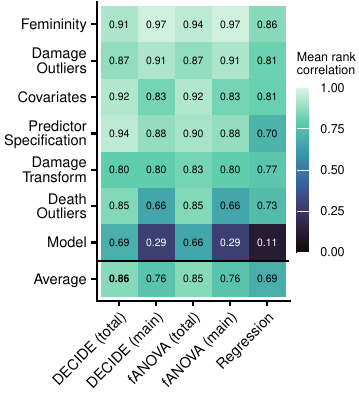}\vspace{0.2in}}
    \caption{Ablation stability on the \textit{Hurricane / Jung} multiverse.
    \textbf{Left:} agreement splits along total vs.\ main effects: the two total-effect estimators agree almost perfectly with each other, as do the two main-effect estimators, while pairs across the two groups agree less, most visibly on the full multiverse. \textbf{Right:} $\method_{\texttt{total}}$ and $\text{fANOVA}_{\texttt{total}}$ are the most stable on average; fixing `Model' destabilizes all other estimators.}
    \label{fig:hurricane-ablation-inter-intra}
\end{figure}

\begin{table}[t]
\setlength{\tabcolsep}{4.5pt}
\caption{\label{tab:importance-hurricane}Importance estimates per decision, by method for the \textit{Hurricane / Jung} multiverse~\citep{simonsohn2020specification,jung2014female}.}
\centering
\begin{tabular}[t]{lrrrrrr}
\toprule
\multicolumn{1}{c}{ } & \multicolumn{3}{c}{\textsc{Decide}} & \multicolumn{2}{c}{fANOVA} & \multicolumn{1}{c}{ } \\
\cmidrule(l{3pt}r{3pt}){2-4} \cmidrule(l{3pt}r{3pt}){5-6}
Decision & Total & Main & Ratio & Total & Main & Reg.\\
\midrule
Model & 2.251 & 1.004 & 44.6\% & 0.540 & 0.247 & 2.004\\
Death Outliers & 1.714 & 0.722 & 42.1\% & 0.427 & 0.175 & 1.006\\
Damage Transform & 1.035 & 0.001 & 0.1\% & 0.231 & 0.000 & 0.059\\
Damage Outliers & 0.964 & 0.157 & 16.3\% & 0.229 & 0.041 & 0.336\\
Covariates & 0.771 & 0.057 & 7.4\% & 0.186 & 0.013 & 0.383\\
Predictor Specification & 0.758 & 0.103 & 13.6\% & 0.177 & 0.021 & 0.561\\
Femininity & 0.261 & 0.034 & 12.9\% & 0.060 & 0.007 & 0.367\\
\bottomrule
\end{tabular}
\end{table}

Another commonly-examined multiverse was introduced by Simonsohn et al. \citep{simonsohn2020specification} when describing Specification Curve Analysis. Building on work by Jung et al. \citep{jung2014female} which claims that hurricanes with female names are deadlier than ones with male names and sparked a debate about how to properly answer this question \citep{malter2014female,bakkensen2014population,maley2014statistics,bakkensen2014population,christensen2014female,jung2014reply}, this multiverse is commonly referred to as \textit{Hurricane}. It consists of the seven decisions displayed in Table~\ref{tab:hurricane-mv-info} of Appendix~\ref{sec:supp-setup}.\footnote{We base the number and definition of decisions on the shared code from Simonsohn et al. \citep{simonsohn2020specification} which does not exactly match with how decisions are described in the main body and supplementary note of the paper.}

Table~\ref{tab:importance-hurricane} reports the full-multiverse importance estimates.
All estimators place `Model' first, but they diverge immediately afterwards.
Most strikingly, $\method_{\texttt{total}}$ and $\text{fANOVA}_{\texttt{total}}$ rank `Damage Transform' third ($1.035$ and $0.231$), while every main-effect estimator places it last with an importance indistinguishable from zero ($\method_{\texttt{main}}$: $0.001$, a ratio of $0.1\%$; $\text{fANOVA}_{\texttt{main}}$: $0.000$; regression: $0.059$).
Whether damages are log-transformed therefore matters only in combination with other decisions, and is invisible to any method that does not account for interactions.
The same pattern holds more mildly for `Covariates' ($7.4\%$) and `Predictor Specification' ($13.6\%$).
We also note that `Femininity'---the decision behind the substantive question---is ranked last by $\method_{\texttt{total}}$, indicating that the reported effect is shaped far more by the surrounding modeling choices than by how the key predictor itself is defined.

Figure~\ref{fig:hurricane-ablation-inter-intra} shows the corresponding inter- and intra-method summaries.
Averaged over the seven ablations, $\method_{\texttt{total}}$ is again the most stable estimator ($\bar{\rho} = 0.86$), followed closely by $\text{fANOVA}_{\texttt{total}}$ ($0.85$), and later $\method_{\texttt{main}}$ and $\text{fANOVA}_{\texttt{main}}$ (both $\approx 0.76$) as well as regression ($0.69$).
Fixing the `Model' decision is the most destabilizing ablation for every estimator, but the estimators degrade very differently: the regression rankings essentially decorrelate from their full-multiverse counterpart ($\bar{\rho} = 0.11$, dropping to $-0.26$ for one of the two choices) and the main-effect estimators fall to $0.29$, whereas the total-effect estimators retain $0.66$--$0.69$.
This is consistent with `Model' carrying much of the interaction structure of this multiverse, which only the total-effect variants account for.

\subsection{Adolescent Well-Being / Orben \& Przybylski}
\label{subsec:orben-wellbeing}

\begin{table}[htbp]
\centering
\setlength{\tabcolsep}{4.5pt}
\caption{Importance estimates per decision, by method for the three multiverses in Orben \& Przybylski \citep{orben2019association}.}
\label{tab:importance-all}

\begin{subtable}[t]{\linewidth}
\centering
\caption{YRBS}
\label{tab:importance-yrbs}
\begin{tabular}[t]{lrrrrrr}
\toprule
\multicolumn{1}{c}{ } & \multicolumn{3}{c}{\textsc{Decide}} & \multicolumn{2}{c}{fANOVA} & \multicolumn{1}{c}{ } \\
\cmidrule(l{3pt}r{3pt}){2-4} \cmidrule(l{3pt}r{3pt}){5-6}
Decision & Total & Main & Ratio & Total & Main & Reg.\\
\midrule
X Variable & 0.00063 & 0.00050 & 78.7\% & 0.86082 & 0.68046 & 0.04572\\
Combination & 0.00019 & 0.00008 & 41.8\% & 0.21577 & 0.06954 & 0.01492\\
Y Variable & 0.00012 & 0.00002 & 18.4\% & 0.21565 & 0.02518 & 0.01847\\
Controls & 0.00000 & 0.00000 & 54.9\% & 0.01152 & 0.00258 & 0.00315\\
\bottomrule
\end{tabular}
\end{subtable}

\vspace{1em}

\begin{subtable}[t]{\linewidth}
\centering
\caption{MTF}
\label{tab:importance-mtf}
\begin{tabular}[t]{lrrrrr}
\toprule
\multicolumn{1}{c}{ } & \multicolumn{3}{c}{\textsc{Decide}} & \multicolumn{1}{c}{fANOVA} & \multicolumn{1}{c}{ } \\
\cmidrule(l{3pt}r{3pt}){2-4} \cmidrule(l{3pt}r{3pt}){5-5}
Decision & Total & Main & Ratio & Main & Reg.\\
\midrule
X Variable & 0.00096 & 0.00077 & 80.4\% & 0.45974 & 0.02940\\
Controls & 0.00018 & 0.00007 & 38.0\% & 0.20429 & 0.01661\\
Y Variable & 0.00003 & 0.00003 & 88.3\% & 0.02634 & 0.02888\\
\bottomrule
\end{tabular}
\end{subtable}

\vspace{1em}

\begin{subtable}[t]{\linewidth}
\centering
\caption{MCS}
\label{tab:importance-mcs}
\begin{tabular}[t]{lrrrrr}
\toprule
\multicolumn{1}{c}{ } & \multicolumn{3}{c}{\textsc{Decide}} & \multicolumn{1}{c}{fANOVA} & \multicolumn{1}{c}{ } \\
\cmidrule(l{3pt}r{3pt}){2-4} \cmidrule(l{3pt}r{3pt}){5-5}
Decision & Total & Main & Ratio & Main & Reg.\\
\midrule
X Variable & 0.00536 & 0.00113 & 21.1\% & 0.46764 & 0.07568\\
Y Variable / Respondent & 0.00436 & 0.00068 & 15.5\% & 0.00314 & 0.21440\\
Controls & 0.00215 & 0.00134 & 62.3\% & 0.29560 & 0.07309\\
\bottomrule
\end{tabular}
\end{subtable}

\end{table}

Last, we examine the multiverse analyses introduced in Orben and Przybylski \citep{orben2019association}, who examine the relationship between adolescent technology use and well-being across several large survey datasets. They present a total of three different multiverse analyses, each built around a different survey dataset: (1) the US Youth Risk and Behavior Survey (\textit{YRBS}) \citep{yrbs}, with four decisions (Table~\ref{tab:yrbs-mv-info}); (2) the US Monitoring the Future study (\textit{MTF}) \citep{mtf}, with three decisions (Table~\ref{tab:mtf-mv-info}); and (3) the UK Millennium Cohort Study (\textit{MCS}) \citep{mcs}, with three\footnote{Technically MCS has four decisions, but one of the decisions, `Respondent', is fully determined by the decision `Y Variable'. We therefore merge the two decisions, as discussed in Section~\ref{subsec:method-sparse}.} decisions (Table~\ref{tab:mcs-mv-info}).
The decision `Y Variable' is a composite of many (plausible) outcome variables, which are one-hot encoded for the regression and fANOVA estimators (neither of which handles a categorical variable with hundreds or thousands of levels well) before aggregating the expanded importances back into a single entry. \method\ operates on the decisions directly. For the two larger multiverses (\textit{MTF} and \textit{MCS}), we omit fANOVA total importance, whose full interaction decomposition does not scale well to this size of multiverse.
One oddity of all three multiverses is, that they vary only $d = 4$ (\textit{YRBS}) or $d = 3$ (\textit{MTF}, \textit{MCS}) decisions. As this severely limits the applicability of rank correlations, we use Pearson correlations between importance vectors for these three multiverses instead of the Spearman correlations used for the earlier sections. For the two multiverses with only $d = 3$ decisions we do not conduct ablations (Section~\ref{subsec:ablation-setup}), as this leaves only $d = 2$ importances after ablation.

Table~\ref{tab:importance-all} reports the full-multiverse importance estimates for each of the three multiverses, and Figure~\ref{fig:orben-ablation-inter-intra} displayes the corresponding results from the ablation study.

For YRBS (Table~\ref{tab:importance-yrbs}) and MTF (Table~\ref{tab:importance-mtf}) methods largely agree: `X Variable', the operationalization of technology use, is ranked as most important by all methods across both datasets. As we can see, through \method\ its influence is largely via a main effect, with ratios of $78.7\%$ (YRBS) and $80.4\%$ (MTF) attributed to main effets rather than interactions. While the linear regression generally agrees with the other two methods, it disambiguates variable importance much less, with `X Variable' and `Y Variable' receiving almost equal scores for MTF. For YRBS, correlations between \method\ and fANOVA scores are also very high at $\geq 0.98$ and $\geq 0.95$ for regression (Figure~\ref{fig:yrbs-ablation-inter-method}). The intra-method stability for YRBS, examined via ablation, is high as well, with regression ($0.94$), $\method_{\texttt{total}}$ ($0.93$) and $\text{fANOVA}_{\texttt{total}}$ ($0.92$) at the top (Figure~\ref{fig:yrbs-ablation-intra-method}).
Methods disagree much more on MCS (Table~\ref{tab:importance-mcs}). While $\method_{\texttt{total}}$  and fANOVA both rank `X Variable' first they disagree on the other decisions, with $\method_{\texttt{total}}$ placing `Y Variable' as a close second while fANOVA assigns it almost no importance at all ($0.003$). Regression on the other hand ranks `Y Variable' as the most important decision. This effect becomes even clearer when examining actual correlations (Figure~\ref{fig:mcs-ablation-inter-method}), with multiple correlations going into the negative, especially between Regression and the other two approaches.
The ratio column indicates why: as most decisions' importances are driven by interactions (main-effect ratios of only $21.1\%$ and $15.5\%$ for `X' and `Y Variable', respectively), fANOVA and regression struggle to attribute variation.

\subsection{Computational Cost}
\label{subsec:computational-cost}

\begin{table}[ht]
\setlength{\tabcolsep}{4.5pt}
\caption{\label{tab:profiling-timings}Median execution time per method and multiverse, over 3 runs. $N_{\mathcal{M}}$ denotes the size of the multiverse. Dashes mark methods not evaluated for a given multiverse due to computational constraints. \method\ includes several checks regarding data structure in its timing.}
\centering
\begin{tabular}[t]{lrrrrr}
\toprule
\multicolumn{3}{c}{ } & \method & \multicolumn{2}{c}{fANOVA} \\
\cmidrule(l{3pt}r{3pt}){5-6}
Multiverse & $N_{\mathcal{M}}$ & Reg. & Main + Total & Main & Total\\
\midrule
\textit{Fertility / Durante} & 210 & 11 ms & 144 ms & 132 ms & 345 ms\\
\textit{Hurricane / Jung} & 1,728 & 6 ms & 220 ms & 189 ms & 7.3 s\\
\textit{YRBS} & 372 & 5 ms & 91 ms & 119 ms & 3.2 s\\
\textit{MTF} & 40,966 & 49 ms & 766 ms & 4.4 s & --\\
\textit{MCS} & 20,004 & 84 ms & 181 ms & 13.9 s & --\\
\bottomrule
\end{tabular}
\end{table}

Table~\ref{tab:profiling-timings} reports the median execution time over 3 runs of each method across the five multiverses.
\method\ computes both main and total importances in well under a second on every multiverse and is substantially faster than fANOVA on the larger multiverses, even when computing main and total effects compared to only main effects for the fANOVA: about six times faster on \textit{MTF} ($766$ms vs. $4.4$s) and about $\sim 75$ times faster on \textit{MCS} ($181$ms vs. $13.9$s).
The gap is widest for $\text{fANOVA}_{\texttt{total}}$, whose full interaction decomposition we were unable to evaluate at all for \textit{MTF} and \textit{MCS}. On the medium-sized \textit{Hurricane / Jung} multiverse, \method\ is $\sim 33$  times faster than $\text{fANOVA}_{\texttt{total}}$ ($220$ms vs. $7.3$s).
On the two smaller multiverses, \method\ is slightly slower than $\text{fANOVA}_{\texttt{main}}$ ($144$ms vs. $132$ms; $220$ms vs. $189$ms).
Regression coefficients are the cheapest importances to obtain, but, as the preceding case studies show, they are also the least stable and cannot exhaustively express interaction structure without a priori knowledge.
Since \method\ fits no surrogate model, its cost is driven by the size of the multiverse rather than by the complexity of a fitted proxy.

\section{Beyond Average-Case Importance}
\label{sec:beyond-average-case}

Sections \ref{sec:model-dependent} and \ref{sec:ablation} instantiate the \method\ framework using the mean ($\Lambda$) and variance ($\Phi$) to capture each decision's average influence.
While this setup benefits from the desirable properties discussed in Section~\ref{sec:framework}, the \method\ framework's versatility stems from the analyst's ability to select these operators to answer entirely different questions.
In this section, we leverage this modularity to evaluate worst-case vulnerability by setting $\Phi$ as the range and $\Lambda$ as the maximum.
The resulting scores capture the maximum extent each decision can alter an outcome rather than its average effect.
This shift is especially useful for outlier-sensitive domains, where we would like to detect decisions that are benign on average but capable of triggering a rare but catastrophic failure.
Surrogate-based approaches are unable to offer this interpretation, since each fixes its aggregation by construction and thus cannot obtain a worst-case analogue.
We demonstrate this capability with a synthetic multiverse where the two perspectives are inverted: the decision ranked least important on average is the one capable of doing the most damage.

\subsection{Experimental Setup}

We construct a synthetic multiverse consisting of twelve binary decisions, $c_1,\dots,c_{12}$, yielding $2^{12}=4096$ universes.
These twelve decisions can be split into three groups:
\begin{enumerate}
    \item \textbf{Routine decisions} ($c_1,\dots,c_5$). These decisions influence the accuracy strictly through their main effect. For $i\in\{1,\dots,5\}$, decision $c_i$ has a main effect of $\beta_i=0.064-0.004\cdot i$.
    \item \textbf{Interacting trigger decisions} ($c_6,\dots,c_{11}$). These decisions influence the accuracy through a combination of their main effect and mutual interaction effect. For $i\in\{6,\dots,11\}$, $c_i$ has a main effect of $0.02$.
    \item \textbf{Catastrophic decision} ($c_{12}$). This decision has no effect unless each trigger decision is set to 1, in which case it causes accuracy to decrease by 0.30.
\end{enumerate}
Specifically, the accuracy of a universe $\mathcal{U}$ is defined by:
\[
    f(\mathcal{U})=0.60 + \sum_{i=1}^5\beta_ic_i+0.02\sum_{j=6}^{11}c_j-0.30\cdot c_{12}\cdot\prod_{j=6}^{11}c_j.
\]
The catastrophic outcome only occurs in 32 of the 4096 universes (0.78\% of the multiverse), taking the pipelines it affects from among the best-performing to the worst.

\subsection{Contrasting Average-Case and Worst-Case Importance}

Table~\ref{tab:worst-case} reports the \method\ scores for the multiverse evaluated through both choices of $\Lambda$ and $\Phi$.
Although each routine and trigger decision is more important than $c_{12}$ on average, $c_{12}$ is the most important in the tail.

It is important to note that neither instantiation is incorrect, but rather each is answering a different question.
The average-case scenario measures how much the decision moves the outcome in a typical universe.
Since $c_{12}$ has no effect in 98.4\% of its comparable sets, its average score is heavily diluted by its rarity.

We see that $c_{12}$ is uniquely first in worst-case importance, rather than tying with the group of triggers $c_6,\dots,c_{11}$.
This is because flipping a trigger off eliminates the possibility of the catastrophe, but also forfeits the trigger's fixed effect of 0.02.
Thus, in practice the trigger decisions would each look like a sensible way to improve model performance---but they construct a trap that $c_{12}$ is responsible for springing.

\begin{table}[tb]
  \centering
  \setlength{\tabcolsep}{3pt}
  \caption{\method\ importance scores on the synthetic multiverse. The catastrophic decision $c_{12}$ ranks last under average-case instantiation but first under worst-case instantiation.}
  \begin{tabular}{llrrrr}
    \toprule
    & & \multicolumn{2}{c}{$\Lambda=$ Mean, $\Phi=$ Variance}
      & \multicolumn{2}{c}{$\Lambda=$ Max, $\Phi=$ Range} \\
    \cmidrule(lr){3-4}\cmidrule(lr){5-6}
    Decision & Group & $\method_\texttt{total}$ & Rank
                     & $\method_\texttt{total}$ & Rank \\
    \midrule
    $c_1$              & Routine       & 0.000900 & 1 & 0.060 & 8      \\
    $c_2$              & Routine       & 0.000784 & 2          & 0.056 & 9      \\
    $c_3$              & Routine       & 0.000676 & 3          & 0.052 & 10     \\
    $c_4$              & Routine       & 0.000576 & 4          & 0.048 & 11     \\
    $c_5$              & Routine       & 0.000484 & 5          & 0.044 & 12     \\
    $c_6$--$c_{11}$    & Trigger       & 0.000405 & 6--11      & 0.280 & 2--7   \\
    $c_{12}$           & Catastrophic  & \textbf{0.000352} & \textbf{12} & \textbf{0.300} & \textbf{1} \\
    \bottomrule
  \end{tabular}
  \label{tab:worst-case}
\end{table}

\section{Discussion}
\label{sec:discussion}

The \method\ framework leverages multiverse structure to quantify the contribution each analytical decision makes to outcome variation.
By comparing universes that differ only in the decisions of interest, and evaluating their outcomes using generalized spread ($\Phi$) and summary ($\Lambda$) functions, it isolates main and interaction effects without relying on proxy models.
\method\ recovers Sobol' indices on full multiverses under its mean-variance instantiation, connecting it to the theory of global sensitivity analysis, while its other instantiations target quantities beyond average-case importances.
Across multiple real-world multiverses, \method\ generally achieves superior intra-method stability and preserves decision importance rankings under ablation more consistently than surrogate-based approaches.
By representing a data analysis pipeline's decisions as a multiverse and leveraging this structure to compute model-free importance scores, \method\ helps practitioners and researchers alike determine which judgment calls most influence their results.

\textbf{Limitations.}
\method's importance scores are conditional on the user-specified decision space.
Thus, omitted decisions or options directly alter the resulting rankings.
\method's handling of sparse multiverses encompasses common forms of sparsity, but more irregular instances may require additional adaptation.
Lastly, the framework uses the complete set of universe outputs, which may be infeasible in settings where each universe is expensive to obtain.

\textbf{Future work.}
First, \method\ can provide insight into the debate over multiverse size by identifying non-influential decisions that can be safely pruned.
Second, in settings where evaluating every universe is infeasible, \method\ scores can be estimated from a designed subset of universes, drawing on experimental design and sampling tools; active sampling could also be guided by importance scores to efficiently prioritize high-impact pipelines.
Third, integrating \method\ scores into agentic data science frameworks can allow automated systems to self-audit their analytical decision-making and flag fragile choices to the user.
Lastly, by treating reasoning choices in a language model's chain-of-thought as decisions, \method\ could identify the concrete steps that lead to specific behaviors, thus informing model steering.

\section*{Acknowledgments}
%
The authors would like to thank Anthony Ozerov for helpful discussions that improved the work.
ZTR is supported by the National Science Foundation Graduate Research Fellowship Program under Grant No. DGE-2146752.
%
This work is supported by the DAAD programme Konrad Zuse Schools of Excellence in Artificial Intelligence, sponsored by the Federal Ministry of Education and Research. It is further supported by the Munich Center for Machine Learning (MCML) and the Simons Institute for the Theory of Computing at the University of California, Berkeley. Funded by the Deutsche Forschungsgemeinschaft (DFG, German Research Foundation) under the National Research Data Infrastructure – NFDI 27/2 - 460037581 - BERD@NFDI.
BY is supported by the Research Council of Norway, TRUST-The Norwegian Centre for Trustworthy AI, project number 357425.
BY also gratefully acknowledges partial support from NSF grant DMS-2413265, NSF grant DMS 2209975, NSF grant DMS-2515767, NSF grant 2023505 on Collaborative Research: Foundations of Data Science Institute (FODSI), and NSF grant MC2378 to the Institute for Artificial Cyber Threat Intelligence and OperatioN (ACTION).
Any opinions, findings, and conclusions or recommendations expressed in this material
are those of the authors and do not necessarily reflect the views of the National Science Foundation.


\bibliographystyle{unsrtnat}
\bibliography{references}

@inproceedings{simson2024one,
  title={One model many scores: Using multiverse analysis to prevent fairness hacking and evaluate the influence of model design decisions},
  author={Simson, Jan and Pfisterer, Florian and Kern, Christoph},
  booktitle={Proceedings of the 2024 ACM Conference on Fairness, Accountability, and Transparency},
  pages={1305--1320},
  year={2024}
}

@article{aczel2026investigating,
  title={Investigating the analytical robustness of the social and behavioural sciences},
  author={Aczel, Balazs and Szaszi, Barnabas and Clelland, Harry T and Kovacs, Marton and Holzmeister, Felix and Van Ravenzwaaij, Don and Schulz-K{\"u}mpel, Hannah and Hoffmann, Sabine and Nilsonne, Gustav and Kosa, Livia and others},
  journal={Nature},
  volume={652},
  number={8108},
  pages={135--142},
  year={2026},
  publisher={Nature Publishing Group UK London}
}

@article{
bakkensen2014population,
author = {Laura A. Bakkensen  and William Larson },
title = {Population matters when modeling hurricane fatalities},
journal = {Proceedings of the National Academy of Sciences},
volume = {111},
number = {50},
pages = {E5331-E5332},
year = {2014},
doi = {10.1073/pnas.1417030111},
URL = {https://www.pnas.org/doi/abs/10.1073/pnas.1417030111},
eprint = {https://www.pnas.org/doi/pdf/10.1073/pnas.1417030111}}

@article{
malter2014female,
author = {Daniel Malter},
title = {Female hurricanes are not deadlier than male hurricanes},
journal = {Proceedings of the National Academy of Sciences},
volume = {111},
number = {34},
pages = {E3496-E3496},
year = {2014},
doi = {10.1073/pnas.1411428111},
URL = {https://www.pnas.org/doi/abs/10.1073/pnas.1411428111},
eprint = {https://www.pnas.org/doi/pdf/10.1073/pnas.1411428111}}

@article{
jung2014female,
author = {Kiju Jung  and Sharon Shavitt  and Madhu Viswanathan  and Joseph M. Hilbe },
title = {Female hurricanes are deadlier than male hurricanes},
journal = {Proceedings of the National Academy of Sciences},
volume = {111},
number = {24},
pages = {8782-8787},
year = {2014},
doi = {10.1073/pnas.1402786111},
URL = {https://www.pnas.org/doi/abs/10.1073/pnas.1402786111},
eprint = {https://www.pnas.org/doi/pdf/10.1073/pnas.1402786111}}

@article{durante2013fluctuating,
  title={The fluctuating female vote: Politics, religion, and the ovulatory cycle},
  author={Durante, Kristina M and Rae, Ashley and Griskevicius, Vladas},
  journal={Psychological Science},
  volume={24},
  number={6},
  pages={1007--1016},
  year={2013},
  publisher={Sage Publications Sage CA: Los Angeles, CA}
}

@article{yu2013stability,
  title={Stability},
  author={Yu, Bin},
  journal={Bernoulli},
  pages={1484--1500},
  year={2013},
  publisher={JSTOR}
}

@article{sobol1993sensitivity,
  title={Sensitivity estimates for nonlinear mathematical models.},
  author={Sobol, Ilya M},
  journal={Math. Model. Comput. Exp.},
  volume={1},
  number={4},
  pages={407--414},
  year={1993}
}

@article{bertran2026aianalysts,
  title = {Many {{AI}} Analysts, One Dataset: {{Navigating}} the Agentic Data Science Multiverse},
  shorttitle = {Many {{AI}} Analysts, One Dataset},
  author = {Bertran, Martin and Fogliato, Riccardo and Wu, Zhiwei Steven},
  year = 2026,
  month = jul,
  journal = {Proceedings of the National Academy of Sciences},
  volume = {123},
  number = {29},
  pages = {e2606495123},
  publisher = {Proceedings of the National Academy of Sciences},
  doi = {10.1073/pnas.2606495123},
  url = {https://www.pnas.org/doi/10.1073/pnas.2606495123},
  urldate = {2026-09-03}
}

@article{short2026multicurious,
  title = {Multicurious: {{A Multidisciplinary Guide}} to {{Multiverse Analysis}}},
  shorttitle = {Multicurious},
  author = {Short, Cassie Ann and Breznau, Nate and Bruntsch, Maria and Burkhardt, Micha and Busch, Niko A. and Cesnaite, Elena and Frank, Maximilian and Gie{\ss}ing, Carsten and Kr{\"a}hmer, Daniel and Kristanto, Daniel and Lonsdorf, Tina and Neuendorf, Claudia and Nguyen, Hung H. V. and Rausch, Manuel and Schmalz, Xenia and Schneck, Andreas and Tabakci, Cem and Hildebrandt, Andrea},
  year = 2026,
  month = apr,
  journal = {Advances in Methods and Practices in Psychological Science},
  volume = {9},
  number = {2},
  pages = {25152459261434881},
  publisher = {SAGE Publications Inc},
  issn = {2515-2459},
  doi = {10.1177/25152459261434881},
  url = {https://doi.org/10.1177/25152459261434881},
  urldate = {2026-09-03},
  langid = {english}
}

@inbook{young2025analytical, place={Cambridge}, series={Analytical Methods for Social Research}, title={Influence Analysis and Scope Conditions}, booktitle={Multiverse Analysis: Computational Methods for Robust Results}, publisher={Cambridge University Press}, author={Young, Cristobal and Cumberworth, Erin}, year={2025}, pages={70–97}, collection={Analytical Methods for Social Research}}

@article{
yu2020veridical,
author = {Bin Yu  and Karl Kumbier },
title = {Veridical data science},
journal = {Proceedings of the National Academy of Sciences},
volume = {117},
number = {8},
pages = {3920-3929},
year = {2020},
doi = {10.1073/pnas.1901326117},
URL = {https://www.pnas.org/doi/abs/10.1073/pnas.1901326117},
eprint = {https://www.pnas.org/doi/pdf/10.1073/pnas.1901326117},
}

@article{rewolinski2026pcs,
    author = {Rewolinski, Zachary T. and Yu, Bin},
    title = {Predictability–Computability–Stability workflow for veridical data science in the age of artificial intelligence},
    journal = {Philosophical Transactions of the Royal Society A: Mathematical, Physical and Engineering Sciences},
    volume = {384},
    number = {2321},
    pages = {20240605},
    year = {2026},
    month = {05},
    issn = {1364-503X},
    doi = {10.1098/rsta.2024.0605},
    url = {https://doi.org/10.1098/rsta.2024.0605},
    eprint = {https://royalsocietypublishing.org/rsta/article-pdf/doi/10.1098/rsta.2024.0605/6141788/rsta.2024.0605.pdf},
}

@book{yu2024veridical,
  title={Veridical data science: The practice of responsible data analysis and decision making},
  author={Yu, Bin and Barter, Rebecca L},
  publisher={MIT Press},
  year={2024}
}

@article{gould2025same,
  title={Same data, different analysts: variation in effect sizes due to analytical decisions in ecology and evolutionary biology},
  author={Gould, Elliot and Fraser, Hannah S and Parker, Timothy H and Nakagawa, Shinichi and Griffith, Simon C and Vesk, Peter A and Fidler, Fiona and Hamilton, Daniel G and Abbey-Lee, Robin N and Abbott, Jessica K and others},
  journal={BMC biology},
  volume={23},
  number={1},
  pages={35},
  year={2025},
  publisher={Springer}
}

@Article{begley2012replication,
author={Begley, C. Glenn
and Ellis, Lee M.},
title={Raise standards for preclinical cancer research},
journal={Nature},
year={2012},
month={Mar},
day={01},
volume={483},
number={7391},
pages={531-533},
issn={1476-4687},
doi={10.1038/483531a},
url={https://doi.org/10.1038/483531a}
}

@article{delgiudice2021travelers,
  title = {A {{Traveler}}'s {{Guide}} to the {{Multiverse}}: {{Promises}}, {{Pitfalls}}, and a {{Framework}} for the {{Evaluation}} of {{Analytic Decisions}}},
  shorttitle = {A {{Traveler}}'s {{Guide}} to the {{Multiverse}}},
  author = {Del Giudice, Marco and Gangestad, Steven W.},
  year = 2021,
  month = jan,
  journal = {Advances in Methods and Practices in Psychological Science},
  volume = {4},
  number = {1},
  pages = {2515245920954925},
  publisher = {SAGE Publications Inc},
  issn = {2515-2459},
  doi = {10.1177/2515245920954925},
  url = {https://doi.org/10.1177/2515245920954925},
  urldate = {2025-06-05},
  langid = {english}
}

@article{auspurg2025robustness,
  title = {Robustness Is Better Assessed with a Few Thoughtful Models than with Billions of Regressions},
  author = {Auspurg, Katrin},
  year = 2025,
  month = oct,
  journal = {Proceedings of the National Academy of Sciences},
  volume = {122},
  number = {43},
  pages = {e2521917122},
  publisher = {Proceedings of the National Academy of Sciences},
  doi = {10.1073/pnas.2521917122},
  url = {https://www.pnas.org/doi/abs/10.1073/pnas.2521917122},
  urldate = {2026-03-01}
}

@article{steegen2016increasing,
  title = {Increasing {{Transparency Through}} a {{Multiverse Analysis}}},
  author = {Steegen, Sara and Tuerlinckx, Francis and Gelman, Andrew and Vanpaemel, Wolf},
  year = 2016,
  month = sep,
  journal = {Perspectives on Psychological Science},
  volume = {11},
  number = {5},
  pages = {702--712},
  publisher = {SAGE Publications Inc},
  issn = {1745-6916},
  doi = {10.1177/1745691616658637},
  url = {https://doi.org/10.1177/1745691616658637},
  urldate = {2022-10-20},
  langid = {english}
}

@article{simonsohn2020specification,
  title = {Specification Curve Analysis},
  author = {Simonsohn, Uri and Simmons, Joseph P. and Nelson, Leif D.},
  year = 2020,
  month = nov,
  journal = {Nature Human Behaviour},
  volume = {4},
  number = {11},
  pages = {1208--1214},
  publisher = {Nature Publishing Group},
  issn = {2397-3374},
  doi = {10.1038/s41562-020-0912-z},
  url = {https://www.nature.com/articles/s41562-020-0912-z},
  urldate = {2023-02-07},
  copyright = {2020 The Author(s), under exclusive licence to Springer Nature Limited},
  langid = {english}
}

@article{simmons2011false,
author = {Joseph P. Simmons and Leif D. Nelson and Uri Simonsohn},
title ={False-Positive Psychology: Undisclosed Flexibility in Data Collection and Analysis Allows Presenting Anything as Significant},
journal = {Psychological Science},
volume = {22},
number = {11},
pages = {1359-1366},
year = {2011},
doi = {10.1177/0956797611417632},
note ={PMID: 22006061},
URL = {https://doi.org/10.1177/0956797611417632},
eprint = {https://doi.org/10.1177/0956797611417632}
}

@article{gelman2014statistical,
  title = {The {{Statistical Crisis}} in {{Science}}},
  author = {Gelman, Andrew and Loken, Eric},
  year = 2014,
  journal = {American Scientist},
  volume = {102},
  number = {6},
  pages = {460},
  publisher = {Sigma XI-The Scientific Research Society}
}

@inproceedings{sarma2024milliways,
  title = {Milliways: {{Taming Multiverses}} through {{Principled Evaluation}} of {{Data Analysis Paths}}},
  shorttitle = {Milliways},
  booktitle = {Proceedings of the 2024 {{CHI Conference}} on {{Human Factors}} in {{Computing Systems}}},
  author = {Sarma, Abhraneel and Hwang, Kyle and Hullman, Jessica and Kay, Matthew},
  year = 2024,
  month = may,
  series = {{{CHI}} '24},
  pages = {1--15},
  publisher = {Association for Computing Machinery},
  address = {New York, NY, USA},
  doi = {10.1145/3613904.3642375},
  url = {https://dl.acm.org/doi/10.1145/3613904.3642375},
  urldate = {2025-05-24},
  isbn = {979-8-4007-0330-0}
}

@article{patel2015assessment,
  title = {Assessment of Vibration of Effects Due to Model Specification Can Demonstrate the Instability of Observational Associations},
  author = {Patel, Chirag J. and Burford, Belinda and Ioannidis, John P.A},
  year = 2015,
  month = sep,
  journal = {Journal of clinical epidemiology},
  volume = {68},
  number = {9},
  pages = {1046--1058},
  issn = {0895-4356},
  doi = {10.1016/j.jclinepi.2015.05.029},
  url = {https://pmc.ncbi.nlm.nih.gov/articles/PMC4555355/},
  urldate = {2025-11-02},
  pmcid = {PMC4555355},
  pmid = {26279400}
}

@misc{rohrer2026whats,
  title = {What's a Multiverse Good for Anyway?},
  author = {Rohrer, Julia M and Hullman, Jessica and Gelman, Andrew},
  year = 2026,
  month = feb,
  number = {37g29\_v1},
  publisher = {PsyArXiv},
  url = {https://osf.io/preprints/psyarxiv/37g29_v1/},
  urldate = {2026-03-01}
}

@article{short2025systematic,
  title = {The {{Systematic Multiverse Analysis Registration Tool}} for Defining Multiverse Analyses},
  author = {Short, Cassie Ann and Inceler, Yusuf Co{\c s}ku and Frank, Maximilian and Hildebrandt, Andrea},
  year = 2025,
  month = oct,
  journal = {Royal Society Open Science},
  volume = {12},
  number = {10},
  pages = {250800},
  issn = {2054-5703},
  doi = {10.1098/rsos.250800},
  url = {https://doi.org/10.1098/rsos.250800},
  urldate = {2026-03-01}
}

@article{hooker2007generalized,
  title = {Generalized {{Functional ANOVA Diagnostics}} for {{High-Dimensional Functions}} of {{Dependent Variables}}},
  author = {Hooker, Giles},
  year = 2007,
  journal = {Journal of Computational and Graphical Statistics},
  volume = {16},
  number = {3},
  eprint = {27594267},
  eprinttype = {jstor},
  pages = {709--732},
  issn = {1061-8600},
  url = {https://www.jstor.org/stable/27594267},
  urldate = {2023-04-03}
}

@inproceedings{hutter2014efficient,
  title = {An {{Efficient Approach}} for {{Assessing Hyperparameter Importance}}},
  booktitle = {Proceedings of the 31st {{International Conference}} on {{Machine Learning}}},
  author = {Hutter, Frank and Hoos, Holger and {Leyton-Brown}, Kevin},
  year = 2014,
  month = jan,
  pages = {754--762},
  publisher = {PMLR},
  issn = {1938-7228},
  url = {https://proceedings.mlr.press/v32/hutter14.html},
  urldate = {2023-02-27},
  langid = {english}
}

@article{silberzahn2015crowdsourced,
  title = {Crowdsourced Research: {{Many}} Hands Make Tight Work},
  shorttitle = {Crowdsourced Research},
  author = {Silberzahn, Raphael and Uhlmann, Eric L.},
  year = 2015,
  month = oct,
  journal = {Nature},
  volume = {526},
  number = {7572},
  pages = {189--191},
  publisher = {Nature Publishing Group},
  issn = {1476-4687},
  doi = {10.1038/526189a},
  url = {https://www.nature.com/articles/526189a},
  urldate = {2025-10-13},
  copyright = {2015 Springer Nature Limited},
  langid = {english}
}

@article{silberzahn2018manya,
  title = {Many {{Analysts}}, {{One Data Set}}: {{Making Transparent How Variations}} in {{Analytic Choices Affect Results}}},
  shorttitle = {Many {{Analysts}}, {{One Data Set}}},
  author = {Silberzahn, R. and Uhlmann, E. L. and Martin, D. P. and Anselmi, P. and Aust, F. and Awtrey, E. and Bahn{\'i}k, {\v S}. and Bai, F. and Bannard, C. and Bonnier, E. and Carlsson, R. and Cheung, F. and Christensen, G. and Clay, R. and Craig, M. A. and Dalla Rosa, A. and Dam, L. and Evans, M. H. and Flores Cervantes, I. and Fong, N. and {Gamez-Djokic}, M. and Glenz, A. and {Gordon-McKeon}, S. and Heaton, T. J. and Hederos, K. and Heene, M. and Hofelich Mohr, A. J. and H{\"o}gden, F. and Hui, K. and Johannesson, M. and Kalodimos, J. and Kaszubowski, E. and Kennedy, D. M. and Lei, R. and Lindsay, T. A. and Liverani, S. and Madan, C. R. and Molden, D. and Molleman, E. and Morey, R. D. and Mulder, L. B. and Nijstad, B. R. and Pope, N. G. and Pope, B. and Prenoveau, J. M. and Rink, F. and Robusto, E. and Roderique, H. and Sandberg, A. and Schl{\"u}ter, E. and Sch{\"o}nbrodt, F. D. and Sherman, M. F. and Sommer, S. A. and Sotak, K. and Spain, S. and Sp{\"o}rlein, C. and Stafford, T. and Stefanutti, L. and Tauber, S. and Ullrich, J. and Vianello, M. and Wagenmakers, E.-J. and Witkowiak, M. and Yoon, S. and Nosek, B. A.},
  year = 2018,
  month = sep,
  journal = {Advances in Methods and Practices in Psychological Science},
  volume = {1},
  number = {3},
  pages = {337--356},
  publisher = {SAGE Publications Inc},
  issn = {2515-2459},
  doi = {10.1177/2515245917747646},
  url = {https://doi.org/10.1177/2515245917747646},
  urldate = {2025-10-13},
  langid = {english}
}

@article{breznau2022observing,
  title = {Observing Many Researchers Using the Same Data and Hypothesis Reveals a Hidden Universe of Uncertainty},
  author = {Breznau, Nate and Rinke, Eike Mark and Wuttke, Alexander and Nguyen, Hung H. V. and Adem, Muna and Adriaans, Jule and {Alvarez-Benjumea}, Amalia and Andersen, Henrik K. and Auer, Daniel and Azevedo, Flavio and Bahnsen, Oke and Balzer, Dave and Bauer, Gerrit and Bauer, Paul C. and Baumann, Markus and Baute, Sharon and Benoit, Verena and Bernauer, Julian and Berning, Carl and Berthold, Anna and Bethke, Felix S. and Biegert, Thomas and Blinzler, Katharina and Blumenberg, Johannes N. and Bobzien, Licia and Bohman, Andrea and Bol, Thijs and Bostic, Amie and Brzozowska, Zuzanna and Burgdorf, Katharina and Burger, Kaspar and Busch, Kathrin B. and {Carlos-Castillo}, Juan and Chan, Nathan and Christmann, Pablo and Connelly, Roxanne and Czymara, Christian S. and Damian, Elena and Ecker, Alejandro and Edelmann, Achim and Eger, Maureen A. and Ellerbrock, Simon and Forke, Anna and Forster, Andrea and Gaasendam, Chris and Gavras, Konstantin and Gayle, Vernon and Gessler, Theresa and Gnambs, Timo and Godefroidt, Am{\'e}lie and Gr{\"o}mping, Max and Gro{\ss}, Martin and Gruber, Stefan and Gummer, Tobias and Hadjar, Andreas and Heisig, Jan Paul and Hellmeier, Sebastian and Heyne, Stefanie and Hirsch, Magdalena and Hjerm, Mikael and Hochman, Oshrat and H{\"o}vermann, Andreas and Hunger, Sophia and Hunkler, Christian and Huth, Nora and Ign{\'a}cz, Zs{\'o}fia S. and Jacobs, Laura and Jacobsen, Jannes and Jaeger, Bastian and Jungkunz, Sebastian and Jungmann, Nils and Kauff, Mathias and Kleinert, Manuel and Klinger, Julia and Kolb, Jan-Philipp and Ko{\l}czy{\'n}ska, Marta and Kuk, John and Kuni{\ss}en, Katharina and Kurti Sinatra, Dafina and Langenkamp, Alexander and Lersch, Philipp M. and L{\"o}bel, Lea-Maria and Lutscher, Philipp and Mader, Matthias and Madia, Joan E. and Malancu, Natalia and Maldonado, Luis and Marahrens, Helge and Martin, Nicole and Martinez, Paul and Mayerl, Jochen and Mayorga, Oscar J. and McManus, Patricia and McWagner, Kyle and Meeusen, Cecil and Meierrieks, Daniel and Mellon, Jonathan and Merhout, Friedolin and Merk, Samuel and Meyer, Daniel and Micheli, Leticia and Mijs, Jonathan and Moya, Crist{\'o}bal and Neunhoeffer, Marcel and N{\"u}st, Daniel and Nyg{\aa}rd, Olav and Ochsenfeld, Fabian and Otte, Gunnar and Pechenkina, Anna O. and Prosser, Christopher and Raes, Louis and Ralston, Kevin and Ramos, Miguel R. and Roets, Arne and Rogers, Jonathan and Ropers, Guido and Samuel, Robin and Sand, Gregor and Schachter, Ariela and Schaeffer, Merlin and Schieferdecker, David and Schlueter, Elmar and Schmidt, Regine and Schmidt, Katja M. and {Schmidt-Catran}, Alexander and Schmiedeberg, Claudia and Schneider, J{\"u}rgen and Schoonvelde, Martijn and {Schulte-Cloos}, Julia and Schumann, Sandy and Schunck, Reinhard and Schupp, J{\"u}rgen and Seuring, Julian and Silber, Henning and Sleegers, Willem and Sonntag, Nico and Staudt, Alexander and Steiber, Nadia and Steiner, Nils and Sternberg, Sebastian and Stiers, Dieter and Stojmenovska, Dragana and Storz, Nora and Striessnig, Erich and Stroppe, Anne-Kathrin and Teltemann, Janna and Tibajev, Andrey and Tung, Brian and Vagni, Giacomo and Van Assche, Jasper and {van der Linden}, Meta and {van der Noll}, Jolanda and Van Hootegem, Arno and Vogtenhuber, Stefan and Voicu, Bogdan and Wagemans, Fieke and Wehl, Nadja and Werner, Hannah and Wiernik, Brenton M. and Winter, Fabian and Wolf, Christof and Yamada, Yuki and Zhang, Nan and Ziller, Conrad and Zins, Stefan and {\.Z}{\'o}{\l}tak, Tomasz},
  year = 2022,
  month = nov,
  journal = {Proceedings of the National Academy of Sciences},
  volume = {119},
  number = {44},
  pages = {e2203150119},
  publisher = {Proceedings of the National Academy of Sciences},
  doi = {10.1073/pnas.2203150119},
  url = {https://www.pnas.org/doi/10.1073/pnas.2203150119},
  urldate = {2022-10-31}
}

@article{coretta2023multidimensional,
  title = {Multidimensional {{Signals}} and {{Analytic Flexibility}}: {{Estimating Degrees}} of {{Freedom}} in {{Human-Speech Analyses}}},
  shorttitle = {Multidimensional {{Signals}} and {{Analytic Flexibility}}},
  author = {Coretta, Stefano and Casillas, Joseph V. and Roessig, Simon and Franke, Michael and Ahn, Byron and {Al-Hoorie}, Ali H. and {Al-Tamimi}, Jalal and Alotaibi, Najd E. and AlShakhori, Mohammed K. and Altmiller, Ruth M. and Arantes, Pablo and Athanasopoulou, Angeliki and {Baese-Berk}, Melissa M. and Bailey, George and Sangma, Cheman Baira A and Beier, Eleonora J. and Benavides, Gabriela M. and Benker, Nicole and BensonMeyer, Emelia P. and Benway, Nina R. and Berry, Grant M. and Bing, Liwen and Bjorndahl, Christina and Bolyanatz, Mari{\v s}ka and Braver, Aaron and Brown, Violet A. and Brown, Alicia M. and Brugos, Alejna and Buchanan, Erin M. and Butlin, Tanna and {Bux{\'o}-Lugo}, Andr{\'e}s and Caillol, Coline and Cangemi, Francesco and Carignan, Christopher and Carraturo, Sita and Caudrelier, Tiphaine and Chodroff, Eleanor and Cohn, Michelle and Cronenberg, Johanna and Crouzet, Olivier and Dagar, Erica L. and Dawson, Charlotte and Diantoro, Carissa A. and Dokovova, Marie and Drake, Shiloh and Du, Fengting and Dubuis, Margaux and Du{\^e}me, Florent and Durward, Matthew and Egurtzegi, Ander and Elsherif, Mahmoud M. and Esser, Janina and Ferragne, Emmanuel and Ferreira, Fernanda and Fink, Lauren K. and Finley, Sara and Foster, Kurtis and Foulkes, Paul and Franzke, Rosa and {Frazer-McKee}, Gabriel and Fromont, Robert and Garc{\'i}a, Christina and Geller, Jason and Grasso, Camille L. and Greca, Pia and Grice, Martine and {Grose-Hodge}, Magdalena S. and Gully, Amelia J. and Halfacre, Caitlin and Hauser, Ivy and Hay, Jen and Haywood, Robert and Hellmuth, Sam and Hilger, Allison I. and Holliday, Nicole and Hoogland, Damar and Huang, Yaqian and Hughes, Vincent and Icardo Isasa, Ane and Ilchovska, Zlatomira G. and Jeon, Hae-Sung and Jones, Jacq and Junges, M{\'a}gat N. and Kaefer, Stephanie and Kaland, Constantijn and Kelley, Matthew C. and Kelly, Niamh E. and Kettig, Thomas and Khattab, Ghada and Koolen, Ruud and Krahmer, Emiel and Krajewska, Dorota and Krug, Andreas and Kumar, Abhilasha A. and Lander, Anna and Lentz, Tomas O. and Li, Wanyin and Li, Yanyu and Lialiou, Maria and LimaJr., Ronaldo M. and Lo, Justin J. H. and Lopez Otero, Julio Cesar and Mackay, Bradley and MacLeod, Bethany and Mallard, Mel and McConnellogue, Carol-Ann Mary and Moroz, George and Murali, Mridhula and Nalborczyk, Ladislas and Nenadi{\'c}, Filip and Nieder, Jessica and Nikoli{\'c}, Du{\v s}an and Nogueira, Francisco G. S. and Offerman, Heather M. and Passoni, Elisa and P{\'e}lissier, Maud and Perry, Scott J. and Pfiffner, Alexandra M. and Proctor, Michael and Rhodes, Ryan and Rodr{\'i}guez, Nicole and Roepke, Elizabeth and R{\"o}er, Jan P. and Sbacco, Lucia and Scarborough, Rebecca and Schaeffler, Felix and Schleef, Erik and Schmitz, Dominic and Shiryaev, Alexander and S{\'o}skuthy, M{\'a}rton and Spaniol, Malin and Stanley, Joseph A. and Strickler, Alyssa and Tavano, Alessandro and Tomaschek, Fabian and Tucker, Benjamin V. and Turnbull, Rory and Ugwuanyi, Kingsley O. and {Urrestarazu-Porta}, I{\~n}igo and {van de Vijver}, Ruben and Van Engen, Kristin J. and {van Miltenburg}, Emiel and Wang, Bruce Xiao and Warner, Natasha and Wehrle, Simon and Westerbeek, Hans and Wiener, Seth and Winters, Stephen and Wong, Sidney G.-J. and Wood, Anna and Wottawa, Jane and Xu, Chenzi and {Z{\'a}rate-S{\'a}ndez}, Germ{\'a}n and Zellou, Georgia and Zhang, Cong and Zhu, Jian and Roettger, Timo B.},
  year = 2023,
  month = jul,
  journal = {Advances in Methods and Practices in Psychological Science},
  volume = {6},
  number = {3},
  pages = {25152459231162567},
  publisher = {SAGE Publications Inc},
  issn = {2515-2459},
  doi = {10.1177/25152459231162567},
  url = {https://doi.org/10.1177/25152459231162567},
  urldate = {2025-11-02},
  langid = {english}
}

@article{ioannidis2005why,
  title = {Why {{Most Published Research Findings Are False}}},
  author = {Ioannidis, John P. A.},
  year = 2005,
  month = aug,
  journal = {PLOS Medicine},
  volume = {2},
  number = {8},
  pages = {e124},
  publisher = {Public Library of Science},
  issn = {1549-1676},
  doi = {10.1371/journal.pmed.0020124},
  url = {https://journals.plos.org/plosmedicine/article?id=10.1371/journal.pmed.0020124},
  urldate = {2025-10-25},
  langid = {english}
}

@article{laraway2019overview,
  title = {An {{Overview}} of {{Scientific Reproducibility}}: {{Consideration}} of {{Relevant Issues}} for {{Behavior Science}}/{{Analysis}}},
  shorttitle = {An {{Overview}} of {{Scientific Reproducibility}}},
  author = {Laraway, Sean and Snycerski, Susan and Pradhan, Sean and Huitema, Bradley E.},
  year = 2019,
  month = mar,
  journal = {Perspectives on Behavior Science},
  volume = {42},
  number = {1},
  pages = {33--57},
  issn = {2520-8977},
  doi = {10.1007/s40614-019-00193-3},
  url = {https://doi.org/10.1007/s40614-019-00193-3},
  urldate = {2025-10-25},
  langid = {english}
}

@article{hutson2018artificiala,
  title = {Artificial Intelligence Faces Reproducibility Crisis},
  author = {Hutson, Matthew},
  year = 2018,
  month = feb,
  journal = {Science},
  volume = {359},
  number = {6377},
  pages = {725--726},
  publisher = {American Association for the Advancement of Science},
  doi = {10.1126/science.359.6377.725},
  url = {https://www.science.org/doi/10.1126/science.359.6377.725},
  urldate = {2025-10-25}
}

@misc{semmelrock2023reproducibility,
  title = {Reproducibility in {{Machine Learning-Driven Research}}},
  author = {Semmelrock, Harald and Kopeinik, Simone and Theiler, Dieter and {Ross-Hellauer}, Tony and Kowald, Dominik},
  year = 2023,
  month = jul,
  number = {arXiv:2307.10320},
  eprint = {2307.10320},
  primaryclass = {cs},
  publisher = {arXiv},
  doi = {10.48550/arXiv.2307.10320},
  url = {http://arxiv.org/abs/2307.10320},
  urldate = {2025-10-25},
  archiveprefix = {arXiv}
}

@article{gundersen2018state,
  title = {State of the {{Art}}: {{Reproducibility}} in {{Artificial Intelligence}}},
  shorttitle = {State of the {{Art}}},
  author = {Gundersen, Odd Erik and Kjensmo, Sigbj{\o}rn},
  year = 2018,
  month = apr,
  journal = {Proceedings of the AAAI Conference on Artificial Intelligence},
  volume = {32},
  number = {1},
  issn = {2374-3468},
  doi = {10.1609/aaai.v32i1.11503},
  url = {https://ojs.aaai.org/index.php/AAAI/article/view/11503},
  urldate = {2025-10-25},
  copyright = {Copyright (c)},
  langid = {english}
}

@inproceedings{herrmann2024position,
  title = {Position: {{Why We Must Rethink Empirical Research}} in {{Machine Learning}}},
  shorttitle = {Position},
  booktitle = {Proceedings of the 41st {{International Conference}} on {{Machine Learning}}},
  author = {Herrmann, Moritz and Lange, F. Julian D. and Eggensperger, Katharina and Casalicchio, Giuseppe and Wever, Marcel and Feurer, Matthias and R{\"u}gamer, David and H{\"u}llermeier, Eyke and Boulesteix, Anne-Laure and Bischl, Bernd},
  year = 2024,
  month = jul,
  pages = {18228--18247},
  publisher = {PMLR},
  issn = {2640-3498},
  url = {https://proceedings.mlr.press/v235/herrmann24b.html},
  urldate = {2025-10-29},
  langid = {english}
}

@article{krahmer2026visualizing,
  title = {Visualizing Vastness: {{Graphical}} Methods for Multiverse Analysis},
  shorttitle = {Visualizing Vastness},
  author = {Kr{\"a}hmer, Daniel and Young, Cristobal},
  year = 2026,
  month = feb,
  journal = {PLOS ONE},
  volume = {21},
  number = {2},
  pages = {e0339452},
  publisher = {Public Library of Science},
  issn = {1932-6203},
  doi = {10.1371/journal.pone.0339452},
  url = {https://journals.plos.org/plosone/article?id=10.1371/journal.pone.0339452},
  urldate = {2026-03-13},
  langid = {english}
}

@article{liu2021boba,
  title = {Boba: {{Authoring}} and {{Visualizing Multiverse Analyses}}},
  shorttitle = {Boba},
  author = {Liu, Yang and Kale, Alex and Althoff, Tim and Heer, Jeffrey},
  year = 2021,
  month = feb,
  journal = {IEEE Transactions on Visualization and Computer Graphics},
  volume = {27},
  number = {2},
  pages = {1753--1763},
  issn = {1941-0506},
  doi = {10.1109/TVCG.2020.3028985},
  urldate = {2025-01-29}
}

@article{orben2019association,
	title = {The association between adolescent well-being and digital technology use},
	volume = {3},
	copyright = {2019 The Author(s), under exclusive licence to Springer Nature Limited},
	issn = {2397-3374},
	url = {https://www.nature.com/articles/s41562-018-0506-1},
	doi = {10.1038/s41562-018-0506-1},
	language = {en},
	number = {2},
	urldate = {2026-05-12},
	journal = {Nature Human Behaviour},
	publisher = {Nature Publishing Group},
	author = {Orben, Amy and Przybylski, Andrew K.},
	month = feb,
	year = {2019},
	pages = {173--182}
}

@misc{yrbs,
  title={Youth risk behavior surveillance—United States, 2015. MMWR Surveillance Summaries, 65 (6), 1--174},
  author={Kann, L and McManus, T and Harris, WA and Shanklin, SL and Flint, KH and Hawkins, J and Queen, B and Lowry, R and Olsen, EO and Chyen, D and others},
  year={2016}
}

@misc{mcs,
  title={Millennium cohort study sixth survey 2015--2016 user guide},
  author={Agalioti-Sgompou, V and Calderwood, L and Gilbert, E and Haselden, L and Johnson, J and Smith, K and Team t, MC},
  year={2017},
  publisher={Centre for longitudinal studies, Institution of Education, University~…}
}

@misc{mtf,
	title = {Monitoring the {Future}: {A} {Continuing} {Study} of {American} {Youth} (8th- and 10th-{Grade} {Surveys}), 2016},
	shorttitle = {Monitoring the {Future}},
	url = {https://www.icpsr.umich.edu/web/NAHDAP/studies/36799/versions/V1},
	doi = {10.3886/ICPSR36799.v1},
	language = {en},
	urldate = {2026-09-01},
	publisher = {Inter-university Consortium for Political and Social Research},
	author = {Johnston, Lloyd D. and Bachman, Jerald G. and O'Malley, Patrick M. and Schulenberg, John E. and Miech, Richard A.},
	month = oct,
	year = {2017},
}

@article{maley2014statistics,
  title={Statistics show no evidence of gender bias in the public's hurricane preparedness},
  author={Maley, Steve},
  journal={Proceedings of the National Academy of Sciences},
  volume={111},
  number={37},
  pages={E3834--E3834},
  year={2014},
  publisher={National Academy of Sciences}
}

@article{christensen2014female,
  title={Are female hurricanes really deadlier than male hurricanes?},
  author={Christensen, Bj{\"o}rn and Christensen, S{\"o}ren},
  journal={Proceedings of the National Academy of Sciences},
  volume={111},
  number={34},
  pages={E3497--E3498},
  year={2014},
  publisher={National Academy of Sciences}
}

@article{jung2014reply,
  title={Reply to Christensen and Christensen and to Malter: Pitfalls of erroneous analyses of hurricanes names},
  author={Jung, Kiju and Shavitt, Sharon and Viswanathan, Madhu and Hilbe, Joseph M},
  journal={Proceedings of the National Academy of Sciences},
  volume={111},
  number={34},
  pages={E3499--E3500},
  year={2014},
  publisher={National Academy of Sciences}
}

@article{homma1996importance,
title = {Importance measures in global sensitivity analysis of nonlinear models},
journal = {Reliability Engineering \& System Safety},
volume = {52},
number = {1},
pages = {1-17},
year = {1996},
issn = {0951-8320},
doi = {https://doi.org/10.1016/0951-8320(96)00002-6},
url = {https://www.sciencedirect.com/science/article/pii/0951832096000026},
author = {Toshimitsu Homma and Andrea Saltelli}
}

@book{saltelli2008global,
  title={Global sensitivity analysis: the primer},
  author={Saltelli, Andrea and Ratto, Marco and Andres, Terry and Campolongo, Francesca and Cariboni, Jessica and Gatelli, Debora and Saisana, Michaela and Tarantola, Stefano},
  year={2008},
  publisher={John Wiley \& Sons}
}

@article{freedman1991statistical,
  title={Statistical models and shoe leather},
  author={Freedman, David A},
  journal={Sociological methodology},
  pages={291--313},
  year={1991},
  publisher={JSTOR}
}

@article{breiman2001statistical,
  title={Statistical modeling: The two cultures (with comments and a rejoinder by the author)},
  author={Breiman, Leo},
  journal={Statistical science},
  volume={16},
  number={3},
  pages={199--231},
  year={2001},
  publisher={Institute of Mathematical Statistics}
}

@book{young2024multiverse,
  title = {Multiverse Analysis: Computational Methods for Robust Results},
  shorttitle = {Multiverse Analysis},
  author = {Young, Cristobal and Cumberworth, Erin},
  year = 2024,
  series = {Analytical Methods for Social Research},
  publisher = {Cambridge University Press},
  address = {New York, NY},
  isbn = {978-1-316-51878-6},
  lccn = {H62 .Y669 2024}
}

@article{cassee2026exploringa,
  title = {Exploring the {{Garden}} of {{Forking Paths}} in {{Empirical Software Engineering Research}}: {{A Multiverse Analysis}}},
  shorttitle = {Exploring the {{Garden}} of {{Forking Paths}} in {{Empirical Software Engineering Research}}},
  author = {Cassee, Nathan and Feldt, Robert},
  year = 2026,
  month = sep,
  journal = {ACM Transactions on Software Engineering and Methodology},
  volume = {35},
  number = {10},
  pages = {323:1--323:26},
  issn = {1049-331X},
  doi = {10.1145/3799711},
  url = {https://dl.acm.org/doi/10.1145/3799711},
  urldate = {2026-09-21}
}

@article{engzell2023understandinga,
  title = {Understanding {{Patterns}} and {{Trends}} in {{Income Mobility}} through {{Multiverse Analysis}}},
  author = {Engzell, Per and Mood, Carina},
  year = 2023,
  month = aug,
  journal = {American Sociological Review},
  volume = {88},
  number = {4},
  pages = {600--626},
  publisher = {SAGE Publications Inc},
  issn = {0003-1224},
  doi = {10.1177/00031224231180607},
  url = {https://doi.org/10.1177/00031224231180607},
  urldate = {2026-09-21},
  langid = {english}
}

@article{ferguson2025uncertaintya,
  title = {Uncertainty in the Estimated Effects of Statin Initiation on Risk of Dementia: Using a Multiverse Analysis to Assess Sources of Variability},
  shorttitle = {Uncertainty in the Estimated Effects of Statin Initiation on Risk of Dementia},
  author = {Ferguson, Erin L. and Zimmerman, Scott C. and Jiang, Chen and Choi, Minhyuk and Meyers, Travis J. and Hoffmann, Thomas J. and Gilsanz, Paola and Wang, Jingxuan and {Oni-Orisan}, Akinyemi and Whitmer, Rachel A. and Risch, Neil and Krauss, Ronald M. and Patel, Chirag J. and Schaefer, Catherine A. and Glymour, M. Maria},
  year = 2025,
  month = apr,
  journal = {European Journal of Epidemiology},
  volume = {40},
  number = {4},
  pages = {407--417},
  issn = {1573-7284},
  doi = {10.1007/s10654-025-01231-y},
  url = {https://doi.org/10.1007/s10654-025-01231-y},
  urldate = {2026-09-21},
  langid = {english}
}

@article{godwin2025multiverse,
  title = {A Multiverse Analysis of Cleaning and Analyzing Procedures of Eye Movement Data during Reading},
  author = {Godwin, Hayward J. and Lee, Charlotte E. and Drieghe, Denis},
  year = 2025,
  month = may,
  journal = {Behavior Research Methods},
  volume = {57},
  number = {6},
  pages = {164},
  issn = {1554-3528},
  doi = {10.3758/s13428-025-02689-0},
  url = {https://link.springer.com/10.3758/s13428-025-02689-0},
  urldate = {2026-09-21},
  langid = {english}
}

@article{plessen2025digital,
  title = {Digital Mental Health Interventions for the Treatment of Depression: {{A}} Multiverse Meta-Analysis},
  shorttitle = {Digital Mental Health Interventions for the Treatment of Depression},
  author = {Plessen, Constantin Yves and Panagiotopoulou, Olga Maria and Tong, Lingyao and Cuijpers, Pim and Karyotaki, Eirini},
  year = 2025,
  month = jan,
  journal = {Journal of Affective Disorders},
  volume = {369},
  pages = {1031--1044},
  issn = {0165-0327},
  doi = {10.1016/j.jad.2024.10.018},
  url = {https://www.sciencedirect.com/science/article/pii/S0165032724016793},
  urldate = {2026-09-21}
}

@inproceedings{saltelli2025global,
  title = {Global {{Sensitivity Analysis Unveils}} the~{{Hidden Universe}} of~{{Uncertainty}} in~{{Multiverse Studies}}},
  booktitle = {Statistics for {{Innovation IV}}},
  author = {Saltelli, Andrea and Lachi, Alessio and Puy, Arnald and Breznau, Nate},
  editor = {{di Bella}, Enrico and Gioia, Vincenzo and Lagazio, Corrado and Zaccarin, Susanna},
  year = 2025,
  pages = {299--304},
  publisher = {Springer Nature Switzerland},
  address = {Cham},
  doi = {10.1007/978-3-031-96033-8_49},
  isbn = {978-3-031-96033-8},
  langid = {english}
}

@article{borgonovo2007new,
title = {A new uncertainty importance measure},
journal = {Reliability Engineering \& System Safety},
volume = {92},
number = {6},
pages = {771-784},
year = {2007},
issn = {0951-8320},
doi = {https://doi.org/10.1016/j.ress.2006.04.015},
url = {https://www.sciencedirect.com/science/article/pii/S0951832006000883},
author = {E. Borgonovo}
}

@article{morris1991factorial,
author = {Max D. Morris},
title = {Factorial Sampling Plans for Preliminary Computational Experiments},
journal = {Technometrics},
volume = {33},
number = {2},
pages = {161--174},
year = {1991},
publisher = {Taylor \& Francis},
doi = {10.1080/00401706.1991.10484804},
URL = {https://www.tandfonline.com/doi/abs/10.1080/00401706.1991.10484804},
eprint = {https://www.tandfonline.com/doi/pdf/10.1080/00401706.1991.10484804}
}

@inproceedings{bell2022modeling,
 author = {Bell, Samuel J. and Kampman, Onno and Dodge, Jesse and Lawrence, Neil},
 booktitle = {Advances in Neural Information Processing Systems},
 doi = {10.52202/068431-1338},
 editor = {S. Koyejo and S. Mohamed and A. Agarwal and D. Belgrave and K. Cho and A. Oh},
 pages = {18416--18429},
 publisher = {Curran Associates, Inc.},
 title = {Modeling the Machine Learning Multiverse},
 url = {https://proceedings.neurips.cc/paper_files/paper/2022/file/750337e1301941f81ae31a90e0a1c181-Paper-Conference.pdf},
 volume = {35},
 year = {2022}
}

@misc{liu2023approximation,
  title = {Approximation and {{Progressive Display}} of {{Multiverse Analyses}}},
  author = {Liu, Yang and Althoff, Tim and Heer, Jeffrey},
  year = 2023,
  publisher = {arXiv},
  doi = {10.48550/ARXIV.2305.08323},
  url = {https://arxiv.org/abs/2305.08323},
  urldate = {2026-09-21},
  copyright = {Creative Commons Attribution 4.0 International}
}
\appendix
\counterwithin*{table}{section}
\counterwithin*{figure}{section}

\renewcommand{\thetable}{\thesection\arabic{table}}
\renewcommand{\thefigure}{\thesection\arabic{figure}}

\clearpage
\section{Supplementary Mathematical Details}

Throughout this section, let $\Lambda$ be the mean and $\Phi$ be the population variance.
Multiverse $\mathcal{M}=[o_1]\times\dots\times[o_d]$ is full and has uniform weight across its universes, although the following arguments also hold whenever the decisions are independent, i.e., whenever universe $\mathcal{U}=(c_1,\dots,c_d)\in\mathcal{M}$ has $\mathbb{P}(\mathcal{U})=\mathbb{P}(c_1)\cdot\dots\cdot\mathbb{P}(c_d)$.
For simplicity of notation across concepts, we let $X=(X_1,\dots,X_d)$ be a universe in $\mathcal{M}$ and say that $Y=f(X)$.
Thus, conditioning on a background $b_{-\mathcal{S}}$ or on choices $c_{\mathcal{S}}$ is equivalent to conditioning on $X_{-\mathcal{S}}$ or $X_{\mathcal{S}}$, respectively.
It follows that the definitions in Section~\ref{subsec:method-main-interaction} can then be written as
\begin{subequations}
\begin{align}
\label{eq:supp-decide-total-defn}
    \method_{\texttt{total}}(\mathcal{S})&=\mathbb{E}\left[\operatorname{Var}\left(Y\mid X_{-\mathcal{S}}\right)\right],\\
\label{eq:supp-decide-main-defn}
    \method_{\texttt{main}}(\mathcal{S})&=\operatorname{Var}\left(\mathbb{E}\left[Y\mid X_{\mathcal{S}}\right]\right).
\end{align}
\end{subequations}

\subsection{Interpretations of Main, Total, and Interaction Effects}
\label{subsec:supp-interp}

The Law of Total Variance tells us that for any random variable $Z$, we have
\begin{equation}
    \operatorname{Var}(Y)=\operatorname{Var}\left(\mathbb{E}[Y\mid Z]\right) + \mathbb{E}\left[\operatorname{Var}(Y\mid Z)\right].
\end{equation}
We take $Z=X_{-\mathcal{S}}$, plug in Equation~\ref{eq:supp-decide-total-defn} to the right-hand side, and divide by $\operatorname{Var}(Y)$ to get
\begin{equation}
\label{eq:interp-decide-total}
    \frac{\method_{\texttt{total}}(\mathcal{S})}{\operatorname{Var}(Y)}=1-\frac{\operatorname{Var}\left(\mathbb{E}[Y\mid X_{-\mathcal{S}}]\right)}{\operatorname{Var}(Y)}.
\end{equation}
The subtracted term on the right-hand side of Equation~\ref{eq:interp-decide-total} represents the share of $\operatorname{Var}(Y)$ that the background solely accounts for.
It follows that $\method_{\texttt{total}}(S)/\operatorname{Var}(Y)$ is the share of $\operatorname{Var}(Y)$ that remains after fixing all decisions other than $S$.

Taking $Z=X_{\mathcal{S}}$ and instead plugging in Equation~\ref{eq:supp-decide-main-defn} gives Equation~\ref{eq:main-effect-decomp}:
\begin{equation}
    \frac{\method_{\texttt{main}}(S)}{\operatorname{Var}(Y)}=1-\frac{\mathbb{E}\left[\operatorname{Var}(Y\mid X_{\mathcal{S}})\right]}{\operatorname{Var}(Y)}.
\end{equation}
Since $\mathbb{E}\left[\operatorname{Var}(Y\mid X_{\mathcal{S}})\right]$ represents the variance that remains after fixing the decisions in $S$, the share of $\operatorname{Var}(Y)$ eliminated by fixing $S$ is thus $\method_{\texttt{main}}(S)/\operatorname{Var}(Y)$.

The ANOVA-HDMR decomposition \citep{saltelli2008global} allows us to write $f$ as a sum of components $f_u$ (one for each subset $u\subseteq[d]$), where $f_u$ depends only on the decisions in $u$.
The decisions in the multiverse are independent, the variances of these components add and we can write
\begin{equation}
\label{eq:variance-sum}
    \operatorname{Var}(Y)=\sum_{u\subseteq[d],u\ne\emptyset}\operatorname{Var}(f_u(X_u)).
\end{equation}
We can combine this with the definition of $\method_{\texttt{total}}$, which tells us that the total effect of decision $i$ splits cleanly into its main effect and higher-order interactions:
\begin{align}
\label{eq:decide-decomp}
    \method_{\texttt{total}}(i)&=\sum_{u:i\in u}\operatorname{Var}(f_u(X_u))\\
    &=\method_{\texttt{main}}(i)+\sum_{u:i\in u, |u|\geq2}\operatorname{Var}(f_u(X_u)).
\end{align}
We then define $\method_{\texttt{interactions}}(i) = \sum_{u:i\in u, |u|\geq2}\operatorname{Var}(f_u(X_u))$, such that
\begin{equation}
    \method_{\texttt{total}}(i)=\method_{\texttt{main}}(i)+\method_{\texttt{interactions}}(i).
\end{equation}
Since $\operatorname{Var}(\cdot)\geq0$, we know that $\method_{\texttt{interactions}}(i)\geq0$, telling us that $\method_{\texttt{total}}(i)\geq\method_{\texttt{main}}(i)$.
Whenever $\method_{\texttt{total}}(i)>0$, the value of $\operatorname{Ratio}_{\texttt{main}}(i)$ from Section~\ref{sec:framework} lies in the unit interval and yields the share of decision $i$'s influence that is due to its main effect.
Furthermore, this decomposition extends to a subset of decisions $S$ by replacing the condition $i\in u$ with $u\cap \mathcal{S}\ne\emptyset$ and $\method_{\texttt{main}}(i)$ with $\sum_{u\subseteq \mathcal{S},u\ne\emptyset}\operatorname{Var}(f_u(X_u))$.
Note that for $|\mathcal{S}|\geq2$ this main effect includes interactions of decisions within $\mathcal{S}$, and thus $\method_{\texttt{interactions}}(\mathcal{S})$ only includes interactions that include at least one decision outside of $\mathcal{S}$.

\subsection{Connection to Sobol' Indices}
\label{subsec:supp-sobol}

The first-order and total-effect Sobol' indices \citep{sobol1993sensitivity, homma1996importance} of an input $X_i$ are defined as
\begin{equation}
    V_i=\frac{\operatorname{Var}\left(\mathbb{E}[Y\mid X_i]\right)}{\operatorname{Var}(Y)}, \quad
    V_{T_i}=\frac{\mathbb{E}[\operatorname{Var}\left(Y\mid X_{-i}\right)]}{\operatorname{Var}(Y)}.
\end{equation}
We can thus plug in the definitions from Equations~\ref{eq:supp-decide-total-defn} and \ref{eq:supp-decide-main-defn} to get
\begin{equation}
    V_i = \frac{\method_{\texttt{main}}(i)}{\operatorname{Var}(Y)}, \quad
    V_{T_i} = \frac{\method_{\texttt{total}}(i)}{\operatorname{Var}(Y)}.
\end{equation}
Applying the same identity for a subset $\mathcal{S}$ with $|\mathcal{S}|\geq2$ rather than an individual decision $i$ yields that $\method_{\texttt{main}}(\mathcal{S})/\operatorname{Var}(Y)$ is the closed Sobol' index of the group and $\method_{\texttt{total}}(\mathcal{S})/\operatorname{Var}(Y)$ is the total Sobol' index of the group \citep{saltelli2008global}.

\subsection{Permutation Feature Importance and Functional ANOVA Under Interpolation}
\label{subsec:supp-pfi-fanova}

Let $\hat{f}$ be a surrogate model that interpolates the multiverse, i.e., $\hat{f}(\mathcal{U})=f(\mathcal{U})$ for all $\mathcal{U}\in\mathcal{M}$.
We show that in this case, permutation feature importance (PFI) and functional ANOVA (fANOVA) importance collapse to \method scores.

\subsubsection{Permutation Feature Importance}

PFI is defined as the increase in mean squared error (MSE) after the column representing the $i$th decision is randomly permuted.
Since $\hat{f}$ interpolates the multiverse, the MSE before the permutation is zero.
After the permutation, each row still represents a universe since $\mathcal{M}$ is complete.
Thus, $\hat{f}$ is equivalent to $f$ on each permuted row.
Recall that since the permutation is uniformly random, the value $X_i'$ of decision $i$ is uniform on $[o_i]$ and independent of the rest of the decisions.
Since $\mathbb{E}\left[(Z-Z')^2\right]=2\operatorname{Var}(Z)$ for i.i.d. $Z$ and $Z'$, we have that
\begin{align}
    \mathbb{E}[\operatorname{PFI}(i)] &= \mathbb{E}\left[\left(f(X_i,X_{-i}) - f(X_i',X_{-i})\right)^2\right]\\
    &= 2\mathbb{E}\left[\operatorname{Var}(Y\mid X_{-i})\right]\\
    &= 2\method_{\texttt{total}}(i).
\end{align}

\subsubsection{Functional ANOVA}

Hutter et al.~\citep{hutter2014efficient} use fANOVA on a random forest to generate hyperparameter importances.
For each tree in the forest, the predictions are marginalized over the decisions outside of each subset $u\subseteq[d]$:
\begin{equation}
\label{eq:fanova-decomp}
    \hat{a}_u(X_u)=\mathbb{E}\left[\hat{f}(X)\mid X_u\right].
\end{equation}
The lower-order terms are then subtracted to obtain the components $\hat{f}_u=\hat{a}_u-\sum_{w\subset u}\hat{f}_w$.
The reported importances are each component's variance $\operatorname{Var}(\hat{f}_u(X_u))$ as a fraction of $\sum_{u\ne\emptyset}\operatorname{Var}(\hat{f}_u(X_u))$, averaged across the trees in the random forest.
The main importance of decision $i$ is then defined as
\begin{equation}
    \frac{\operatorname{Var}(\hat{f}_i(X_i))}{\sum_{u\ne\emptyset}\operatorname{Var}(\hat{f}_u(X_u))},
\end{equation}
and the total importance is
\begin{equation}
    \frac{\sum_{u:i\in u}\operatorname{Var}(\hat{f}_u(X_u))}{\sum_{u\ne\emptyset}\operatorname{Var}(\hat{f}_u(X_u))}.
\end{equation}
Observe that this is the same variance decomposition used for \method\ in Equation~\ref{eq:decide-decomp}, but with the forest predictions $\hat{f}$ being used instead of the ground-truth multiverse output $f$.
Since each decision is a categorical variable, Equation~\ref{eq:fanova-decomp} marginalizes over the universes in $\mathcal{M}$ where $\hat{f}=f$ due to interpolation, giving that $\hat{a}_u=\mathbb{E}[Y\mid X_u]$ and thus $\hat{f}_u=f_u$ for every $u$.
This, together with the fact that $\sum_{u\ne\emptyset}\operatorname{Var}(\hat{f}_u(X_u))=\operatorname{Var}(\hat{f}(X))$ by Equation~\ref{eq:variance-sum}, causes the fANOVA main and total importances to yield:
\begin{align}
    \frac{\operatorname{Var}(\hat{f}_i(X_i))}{\sum_{u\ne\emptyset}\operatorname{Var}(\hat{f}_u(X_u))}&=\frac{\method_{\texttt{main}}(i)}{\operatorname{Var}(Y)},\\
    \frac{\sum_{u:i\in u}\operatorname{Var}(\hat{f}_u(X_u))}{\sum_{u\ne\emptyset}\operatorname{Var}(\hat{f}_u(X_u))}&=\frac{\method_{\texttt{total}}(i)}{\operatorname{Var}(Y)}.
\end{align}
\section{Supplementary Figures}
\label{sec:supp-figures}

\subsection{Framework Visualizations}

\begin{figure*}[htb]
\centering
\begin{subfigure}[b]{0.48\textwidth}
\centering
\begin{tikzpicture}[
  edge from parent/.style={draw, -latex, thin},
  level distance=0.85cm,
  level 1/.style={sibling distance=2.2cm},
  level 2/.style={sibling distance=1.08cm},
  level 3/.style={sibling distance=1.08cm},
  level 4/.style={sibling distance=1.08cm},
  ]
  \node[setupnode] {$\varnothing$}
  child { node[setupnodehigh] {$c_1{=}1$}
    child { node[mergednode] {$c_2{=}a$}
      child { node[setupnode] {$c_3{=}1$} child { node[setupleaf, lnode1] (a1) {$f$} } }
      child { node[setupnode] {$c_3{=}2$} child { node[setupleaf, lnode2] (a2) {$f$} } }
    }
  }
  child { node[setupnodehigh] {$c_1{=}2$}
    child { node[mergednode] {$c_2{=}b$}
      child { node[setupnode] {$c_3{=}1$} child { node[setupleaf, lnode1] (a3) {$f$} } }
      child { node[setupnode] {$c_3{=}2$} child { node[setupleaf, lnode2] (a4) {$f$} } }
    }
  };
  \draw[mvseta, line width=1.6pt, bend right=28] (a1.south) to (a3.south);
  \draw[mvsetb, line width=1.6pt, bend right=28] (a2.south) to (a4.south);
\end{tikzpicture}
\caption{Sparsity through functional dependence. Since $c_2$ is determined by $c_1$, the two are merged into one decision.}
\label{subfig:sparsity-functional-dependence}
\end{subfigure}\hfill
\begin{subfigure}[b]{0.48\textwidth}
\centering
\begin{tikzpicture}[
  edge from parent/.style={draw, -latex, thin},
  level distance=0.85cm,
  level 1/.style={sibling distance=2.6cm},
  level 2/.style={sibling distance=1.25cm},
  level 3/.style={sibling distance=1.25cm},
  ]
  \node[setupnode] {$\varnothing$}
  child { node[setupnode] {$c_1{=}1$}
    child { node[setupnodehigh] {$c_2{=}1$} child { node[noopleaf] (b1) {$f$} } }
    child { node[setupnodehigh] {$c_2{=}2$} child { node[noopleaf] (b2) {$f$} } }
  }
  child { node[setupnode] {$c_1{=}2$}
    child { node[setupnodehigh] {$c_2{=}1$} child { node[setupleaf, lnode1] (b3) {$f$} } }
    child { node[setupnodehigh] {$c_2{=}2$} child { node[setupleaf, lnode1] (b4) {$f$} } }
  };
  \draw[black!35, line width=1.6pt, densely dashed, bend right=28] (b1.south) to node[below, font=\scriptsize, black!50] {no-op} (b2.south);
  \draw[mvseta, line width=1.6pt, bend right=28] (b3.south) to (b4.south);
\end{tikzpicture}
\caption{Sparsity through partially-applicable decisions. Note that $c_2$ has no effect when $c_1{=}1$, so its importance uses only $c_1{=}2$ (applies in 1/2 of backgrounds).}
\label{subfig:sparsity-partially-applicable}
\end{subfigure}

\vspace{0.6em}
\begin{subfigure}[b]{\textwidth}
\centering
\begin{tikzpicture}[
  edge from parent/.style={draw, -latex, thin},
  level distance=0.85cm,
  level 1/.style={sibling distance=5.8cm},
  level 2/.style={sibling distance=2.8cm},
  level 3/.style={sibling distance=1.15cm},
  level 4/.style={sibling distance=1.15cm},
  ]
  \node[setupnode] {$\varnothing$}
  child { node[setupnode] {$c_1{=}1$}
    child { node[setupnodehigh] {$c_2{=}1$}
      child { node[setupnode] {$c_3{=}0$} child { node[setupleaf, lnode1] (c0) {$f$} } }
      child { node[setupnode] {$c_3{=}1$} child { node[setupleaf, lnode2] (c1) {$f$} } }
      child { node[setupnode] {$c_3{=}2$} child { node[setupleaf, lnode3] (c2) {$f$} } }
    }
    child { node[setupnodehigh] {$c_2{=}2$}
      child { node[setupnode] {$c_3{=}1$} child { node[setupleaf, lnode2] (c3) {$f$} } }
      child { node[setupnode] {$c_3{=}2$} child { node[setupleaf, lnode3] (c4) {$f$} } }
    }
  }
  child { node[setupnode] {$c_1{=}2$}
    child { node[setupnodehigh] {$c_2{=}1$}
      child { node[setupnode] {$c_3{=}1$} child { node[setupleaf, lnode4] (c5) {$f$} } }
      child { node[setupnode] {$c_3{=}2$} child { node[setupleaf, draw=mvseta!60!mvsetd, line width=1.6pt] (c6) {$f$} } }
    }
    child { node[setupnodehigh] {$c_2{=}2$}
      child { node[setupnode] {$c_3{=}1$} child { node[setupleaf, lnode4] (c7) {$f$} } }
      child { node[setupnode] {$c_3{=}2$} child { node[setupleaf, draw=mvseta!60!mvsetd, line width=1.6pt] (c8) {$f$} } }
    }
  };
  \node[cfleaf, draw=mvseta, text=mvseta, line width=1.6pt] (cf) at ($(c3)!0.5!(c4) + (0,-1.1)$) {$\bar f$};
  \draw[matchedarrow] (c3.south) to (cf);
  \draw[matchedarrow] (c4.south) to (cf);
  \draw[mvseta, line width=1.6pt, densely dashed, bend right=22] (c0.south) to (cf.west);
  \draw[mvsetb, line width=1.6pt, bend right=28] (c1.south) to (c3.south);
  \draw[mvsetc, line width=1.6pt, bend right=28] (c2.south) to (c4.south);
  \draw[mvsetd, line width=1.6pt, bend right=28] (c5.south) to (c7.south);
  \draw[mvseta!60!mvsetd, line width=1.6pt, bend right=28] (c6.south) to (c8.south);
\end{tikzpicture}
\caption{Sparsity through background-dependent options. Only under $(c_1{=}1, c_2{=}1)$ is $c_3{=}0$ possible. When calculating the importance of $c_2$, its missing counterpart is the matched average $\bar f$ of $(c_1=1, c_2=2, c_3=\cdot)$.}
\label{subfig:sparsity-background-dependent}
\end{subfigure}
\caption{Visualizations of the three types of sparsity described in Section~\ref{subsec:method-sparse}. The decision of interest is denoted in \focustext{orange}. Colored arcs represent comparable sets.}
\label{fig:sparse-types}
\end{figure*}
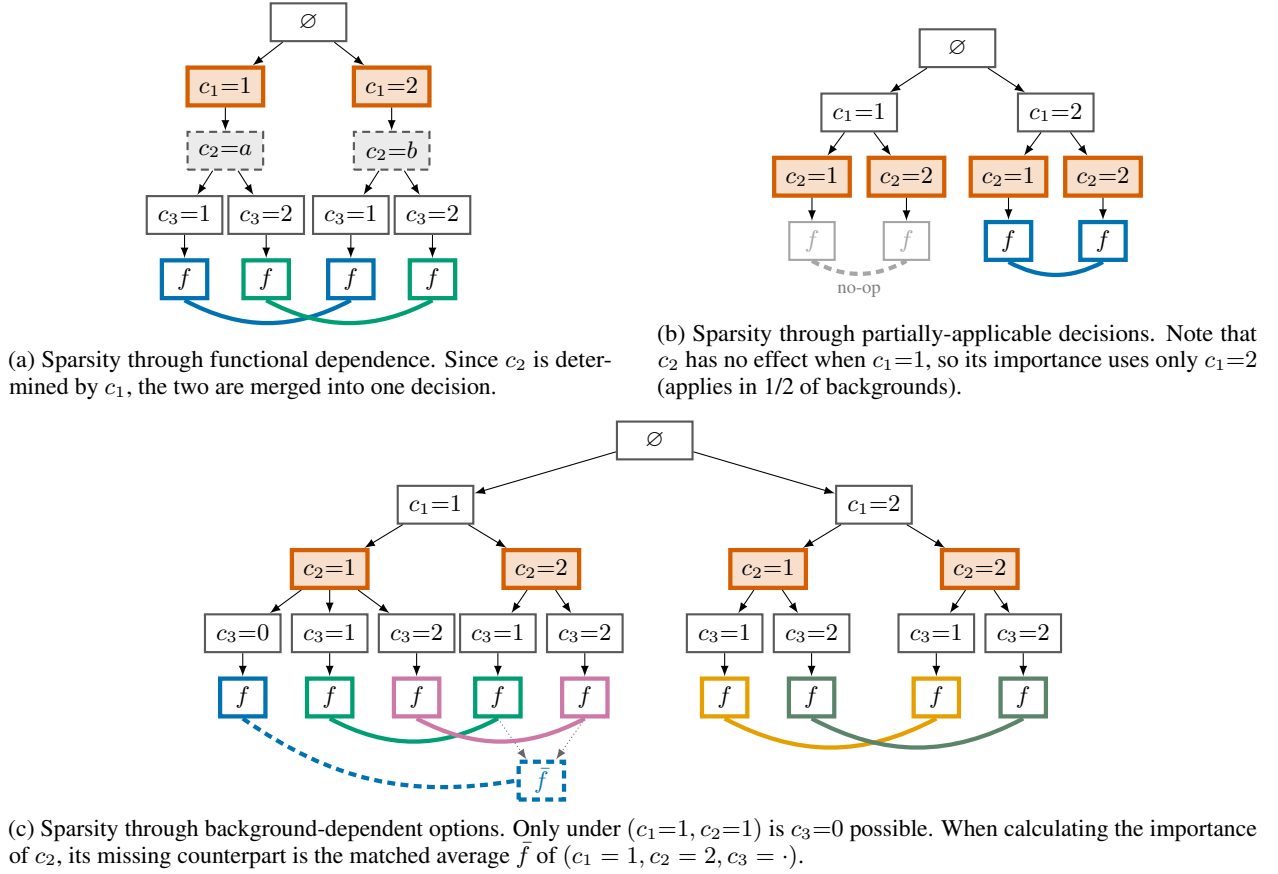

Figure~\ref{fig:sparse-types} displays examples of the three types of sparsity described in Section~\ref{subsec:method-sparse}.
We note that these multiverses are meant to be simple examples, and that real-world multiverses exhibiting these types of sparsity will be larger and more complex in scale.

\subsection{Experimental Result Visualizations}

\begin{figure*}[htb]
    \centering
    \subcaptionbox{YRBS, inter-method stability: Pearson correlation between pairs of methods on the full multiverse (top-right) and averaged over all ablations (bottom-left).\label{fig:yrbs-ablation-inter-method}}
        [3.84in]{\includegraphics[width=\linewidth]{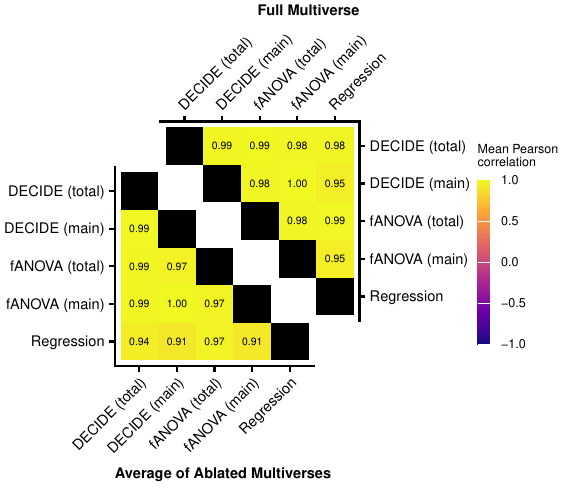}}
    \hfill
    \subcaptionbox{YRBS, intra-method stability: mean Pearson correlation between each method's full-multiverse and ablated importances.\label{fig:yrbs-ablation-intra-method}}
        [2.56in]{\includegraphics[width=\linewidth]{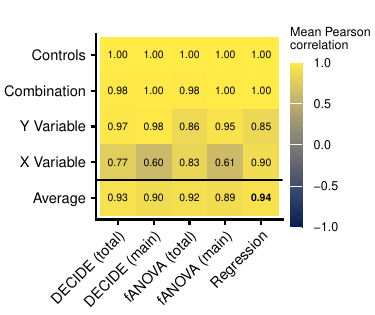}\vspace{0.2in}}

    \vspace{0.5em}
    \subcaptionbox{MTF, inter-method stability: Pearson correlation between pairs of methods on the full multiverse.\label{fig:mtf-ablation-inter-method}}
        [2.85in]{\includegraphics[width=0.9\linewidth]{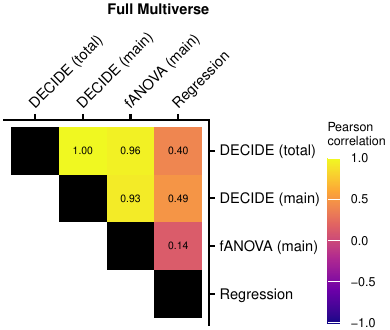}}
    \hfill
    \subcaptionbox{MCS, inter-method stability: Pearson correlation between pairs of methods on the full multiverse.\label{fig:mcs-ablation-inter-method}}
        [2.85in]{\includegraphics[width=0.9\linewidth]{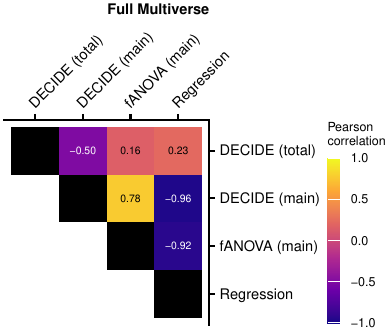}}
    \caption{Method stability for the three adolescent well-being multiverses of Orben and Przybylski \citep{orben2019association}. Top: inter- and intra-method stability under ablation for YRBS. Bottom: inter-method agreement on the full multiverse for MTF and MCS, which vary only three decisions and are therefore not ablated.
    \textbf{YRBS:} every estimator stays highly stable; the most difficult ablation is fixing `X Variable', where the main-effect estimators drop.
    \textbf{MTF:} \method\ and fANOVA agree closely ($0.93$--$1.00$), while regression is only weakly correlated with either ($0.14$--$0.49$).
    \textbf{MCS:} methods disagree strongly; regression is almost perfectly anti-correlated with $\method_{\texttt{main}}$ ($-0.96$) and $\text{fANOVA}_{\texttt{main}}$ ($-0.92$).
    Note that all three multiverses vary at most four judgment calls, so the individual cells are coarse (see the caveat in Section~\ref{subsec:orben-wellbeing}).}
    \label{fig:orben-ablation-inter-intra}
\end{figure*}

Figure~\ref{fig:orben-ablation-inter-intra} shows results from the stability analysis for the three multiverses from Orben \& Przybylski \citep{orben2019association} analyzed in Section~\ref{subsec:orben-wellbeing}.
For YRBS, we show both inter- and intra-method stability under ablation; for MTF and MCS, which vary $d = 3$ decisions, we show inter-method agreement on the full multiverse only, as there are too few decisions to perform meaningful ablations.

\FloatBarrier 
\section{Supplementary Case Study Setup}
\label{sec:supp-setup}

Tables~\ref{tab:fertility-mv-info}-~\ref{tab:mcs-mv-info} display metadata corresponding to the five real-world multiverses used in Section~\ref{sec:ablation}.
We note that \textit{Hurricane / Jung} (Table~\ref{tab:hurricane-mv-info}) and \textit{YRBS} (Table~\ref{tab:yrbs-mv-info}) are full multiverses (i.e., span the entire decision space), whereas \textit{fertility/Durante} (Table~\ref{tab:fertility-mv-info}), \textit{MTF} (Table~\ref{tab:mtf-mv-info}), and \textit{MCS} (Table~\ref{tab:mcs-mv-info}) are sparse multiverses.
In particular, \textit{fertility/Durante} exhibits sparsity through background-dependent options, since valid options for `Menstrual Calculation' depend on the choice of `Cycle Length'.
\textit{MTF} is sparse due to partially-applicable decisions, since `Y Variable' only varies for 5 of the 13 options of `X Variable'.
Lastly, \textit{MCS} has two functionally-dependent choices: `Respondent' is fully determined by `Y Variable', and after merging these decisions the multiverse is otherwise full.

\begin{table}[htbp]
    \centering
    \caption{Decisions in the \textit{Fertility} / \textit{Durante} multiverse. These five decisions lead to $n_{fert} = 210$ universes (less than the theoretical $N_{fert} = 270$ universes due to constraints).}
    \begin{tabular}{@{}l p{0.6\linewidth} r@{}}
        \toprule
        \textbf{Decision} & \textbf{Description} & \textbf{No. of Opts.} \\
        \midrule
        Fertile & Assessment of fertility (high/low) using different windows as cutoffs & 5 \\
        \addlinespace
        Menstrual Calculation & How the next menstrual onset is estimated & 3 \\
        \addlinespace
        Relationship Status & How relationship status is coded, in particular the option ``dating or involved with only one partner'' & 3 \\
        \addlinespace
        Cycle Length & Exclusion criteria for women based on cycle length & 3 \\
        \addlinespace
        Certainty & Exclusion criteria for women based on certainty of start days of previous menstrual onsets & 2 \\
        \bottomrule
    \end{tabular}
    \label{tab:fertility-mv-info}
\end{table}

\begin{table}[htbp]
    \centering
    \caption{Decisions in the \textit{Hurricane} / \textit{Jung} multiverse. These seven decisions lead to $N_{hurr} = 1{,}728$ universes.}
    \begin{tabular}{@{}l p{0.6\linewidth} r@{}}
        \toprule
        \textbf{Decision} & \textbf{Description} & \textbf{No. of Opts.} \\
        \midrule
        Death Outliers & Whether to exclude the storms with the most deaths & 3 \\
        \addlinespace
        Damage Outliers & Whether to exclude the storms with the most property damage & 4 \\
        \addlinespace
        Femininity & Whether the `femininity' of the hurricane name is encoded as continuous or binary & 2 \\
        \addlinespace
        Model & Which model class to use. & 2 \\
        \addlinespace
        Damage Transform & Whether to log-transform damage (i.e., measure with dollars or log-dollars) & 2 \\
        \addlinespace
        Predictor Specification & Whether the effect of `femininity' interacted with other features (and if so, which ones) & 6 \\
        \addlinespace
        Covariates & Which covariates to control for & 3 \\
        \bottomrule
    \end{tabular}
    \label{tab:hurricane-mv-info}
\end{table}

\begin{table}[htbp]
    \centering
    \caption{Decisions in the \textit{YRBS} multiverse. These four decisions lead to $N_{yrbs} = 372$ universes.}
    \begin{tabular}{@{}l p{0.6\linewidth} r@{}}
        \toprule
        \textbf{Decision} & \textbf{Description} & \textbf{No. of Opts.} \\
        \midrule
        X Variable & Operationalization of technology use & 3 \\
        \addlinespace
        Y Variable & Operationalization of well-being & 31 \\
        \addlinespace
        Controls & Which covariates to include & 2 \\
        \addlinespace
        Combination & How well-being measures are combined & 2 \\
        \bottomrule
    \end{tabular}
    \label{tab:yrbs-mv-info}
\end{table}

\begin{table}[htbp]
    \centering
    \caption{Decisions in the \textit{MTF} multiverse. These three decisions lead to $N_{MTF} = 40{,}966$ universes.}
    \begin{tabular}{@{}l p{0.6\linewidth} r@{}}
        \toprule
        \textbf{Decision} & \textbf{Description} & \textbf{No. of Opts.} \\
        \midrule
        X Variable & Operationalization of technology use & 13 \\
        \addlinespace
        Y Variable & Operationalization of well-being & 4{,}095 \\
        \addlinespace
        Controls & Which covariates to include & 2 \\
        \bottomrule
    \end{tabular}
    \label{tab:mtf-mv-info}
\end{table}

\begin{table}[htbp]
    \centering
    \caption{Decisions in the \textit{MCS} multiverse. These four decisions lead to $N_{MCS} = 20{,}004$ universes. The authors selected and analyzed a subset of $n_{MCS} = 20{,}004$ universes out of a theoretical total of $N_{MCS} = 603{,}979{,}752$ defensible universes (or $>2.5$ trillion if any combination of covariates were to be explored). `Respondent' is determined completely by `Y Variable', and is thus merged in accordance with the strategy described in Section~\ref{subsec:method-sparse}.}
    \begin{tabular}{@{}l p{0.6\linewidth} r@{}}
        \toprule
        \textbf{Decision} & \textbf{Description} & \textbf{No. of Opts.} \\
        \midrule
        X Variable & Operationalization of technology use & 6 \\
        \addlinespace
        Y Variable & Operationalization of well-being & 1{,}667 \\
        \addlinespace
        Controls & Which covariates to include & 2 \\
        \addlinespace
        Respondent & Whether to use well-being measures by cohort members or caregivers & 2 \\
        \bottomrule
    \end{tabular}
    \label{tab:mcs-mv-info}
\end{table}

\end{document}